\documentclass[10pt,twocolumn,letterpaper]{article}
\usepackage[pagenumbers]{wacv} 

\usepackage{float}

\definecolor{wacvblue}{rgb}{0.21,0.49,0.74}
\usepackage[pagebackref,breaklinks,colorlinks,allcolors=wacvblue]{hyperref}

\newcommand{\myparagraph}[1]{\noindent{\bf #1}}
\usepackage{multirow} 
\usepackage{colortbl}
\definecolor{oursbg}{RGB}{243,230,217}
\usepackage{algorithm}
\usepackage{algorithmic}

\title{TeMo: \underline{Te}mperature \underline{Mo}dulation for Multimodal Contrastive Learning}

\author{
Dhimitrios Duka$^{1}$ \quad
Bernt Schiele$^{1}$ \quad
Hilde Kuehne$^{2, 3}$ \quad
Anna Kukleva$^{1}$\\
\vspace{0.3em}
{\small $^{1}$MPI for Informatics, SIC \quad
$^{2}$Tuebingen AI Center/University of Tuebingen \quad 
$^{3}$MIT-IBM Watson AI Lab}
}

\begin{document}
\maketitle
\begin{abstract}
Contrastive learning approaches achieve strong performance by training models to bring similar samples closer while pushing dissimilar samples apart. A crucial component of contrastive learning is the temperature hyperparameter $\tau$, which controls the penalty strength applied to negative samples. However, most existing methods either fix this hyperparameter or learn a global value during training. In this paper, we introduce TeMo, \underline{Te}mperature \underline{Mo}dulation framework, a similarity-based modulation approach that adaptively adjusts the temperature for each positive-negative pair according to their similarity, enabling more fine-grained multimodal contrastive learning. Our approach seamlessly integrates temperature-modulated multimodal and unimodal losses with the standard multimodal contrastive loss by gradually transitioning between them. This design allows the model to capture both coarse- and fine-grained semantics at different training stages. Extensive experiments demonstrate that each component of TeMo consistently enhances performance across diverse zero-shot retrieval and classification tasks, establishing new state-of-the-art results.
\end{abstract}
    
\section{Introduction}
Contrastive Learning (CL) has become one of the most relevant self-supervised representation learning paradigms, allowing the training of high performing models on large amounts of unlabeled data. It facilitates the learning of robust unimodal~\cite{he2020momentum, chen2020simple} and multimodal~\cite{radford2021learning, Cherti_2023_CVPR, zhai2023sigmoid, xu2024demystifying} representations by encouraging matching pairs to be close in the embedding space, while pushing nonmatching pairs farther apart.  One of the most important parameters that affect the structure of the learned representations is the temperature $\tau$.
In unimodal learning, $\tau$ plays a key role in controlling the spread of
concepts~\cite{wang2021understanding, kukleva2023temperature}: a higher
temperature tolerates semantically similar samples and forms coarser,
group-wise clusters, whereas a lower temperature sharpens the objective into a
more uniform, instance-level representation space that preserves fine-grained
per-sample detail.
In multimodal learning, however, $\tau$ mainly controls the so-called modality gap~\cite{liang2022mind, wang2021understanding} that separates the vision and language modalities. 
Varying the temperature directly affects the gap: a high temperature closes it and can even eliminate it, whereas a lower temperature drives the modalities farther apart~\cite{liang2022mind}. 
Whereas in this work, we integrate both multimodal and unimodal losses within a unified temperature modulation framework, enabling fine-grained control over the learning dynamics of both multimodal and unimodal representations. We show that this joint modulation not only enhances alignment between modalities by reducing the modality gap, but also improves the structure of unimodal embeddings.

Most of the existing methods~\cite{mu2021slip, radford2021learning, Tang_2025_ICCV, he2020momentum, chen2020improved} use a global learnable temperature parameter shared across all pairs. 
Consequently, the model seeks an overall balance among different semantic classes by applying similar repulsive forces to diverse negative samples.
However, in certain scenarios, such as long-tail datasets, it is more desirable to allow fine-grained control over local structures and underrepresented classes in the embedding space~\cite{kukleva2023temperature}.
Building on this, variable temperature schemes have been proposed to improve representation learning based on temperature alternations~\cite{wang2021understanding, qiu2024cool, kim2026temperature, li2023curriculum, wang2020contextual, zhang2021temperature, khaertdinov2022dynamic, kukleva2023temperature}.

To this end, we propose \underline{Te}mperature \underline{Mo}dulation (TeMo), a novel multimodal contrastive learning framework that leverages combined multimodal and unimodal temperature modulation 
to enhance multimodal representation learning. Unlike prior methods that use a globally assigned temperature~\cite{kukleva2023temperature, qiu2024cool, wang2021understanding}, TeMo adaptively modulates the temperature for \textit{each} positive-negative pair individually. The per-pair temperature strategy enables more precise control over the contrastive objective, allowing the model to dynamically adjust its learning signals based on the similarity of individual paired samples. Additionally, temperature-modulated unimodal losses improve local structure within each modality. We integrate the modulated multimodal and unimodal contrastive losses into a standard CL framework through progressive scheduling. This scheduling enables the model to initially capture instance-level semantic details using a lower temperature, while our adaptive temperature modulation gradually guides sample representations toward coarser semantic groupings.

We evaluate {TeMo} on two standard tasks: multimodal zero-shot retrieval on
MSCOCO and Flickr30k, and zero-shot classification across a diverse set of
datasets, including CIFAR10, CIFAR100, and ImageNet-1k, with models pretrained
on CC3M or CC12M. Across tasks and datasets, {TeMo} consistently outperforms
existing temperature-adaptation baselines. We summarize our contributions as
follows:

\begin{itemize}
    \item We propose {TeMo}, a novel temperature modulation framework for multimodal contrastive learning, which introduces a \textit{per pair} temperature modulation based on the similarity of each pair for both unimodal and multimodal losses, enabling more precise control over the learned representation space;
    \item We show that combining multimodal and unimodal contrastive losses is particularly effective when used with our temperature modulation approach;
    \item We provide an in-depth evaluation of the characteristics of the proposed system and show that {TeMo} outperforms prior temperature adaptation approaches on zero-shot retrieval and classification benchmarks.
\end{itemize}
\section{Related Work}

\myparagraph{Unimodal CL} learns robust representations from single modalities (e.g., images or text) by aligning augmented views of the same input using the InfoNCE loss~\cite{oord2018representation}. Prominent methods include MoCo~\cite{he2020momentum,chen2020improved}, which maintains a momentum-based queue for negatives, and SimCLR~\cite{chen2020simple}, which utilizes samples from the same training batch. The performance of unimodal CL depends on the number of negative samples~\cite{NEURIPS2020_f7cade80,zhang2022dual,yeh2022decoupled}. In our approach, we adopt the batch-wise negative sampling strategy of SimCLR.

\myparagraph{Multimodal CL} extends contrastive learning to multiple modalities, such as images and texts, aiming to align corresponding pairs and distinguish them from unrelated ones. Methods like CLIP~\cite{radford2021learning} optimize a cross-modal InfoNCE objective and demonstrate remarkable generalization in downstream tasks such as zero-shot retrieval and classification. Recent work proposes a wide range of enhancements: DeCLIP~\cite{li2021supervision} and SLIP~\cite{mu2021slip} incorporate vision-specific unimodal self-supervision; CWCL \cite{srinivasa2023cwcl} proposes a new loss function that uses continuous, rather than binary, similarity scores; FILIP~\cite{yao2021filip} and DeFILIP~\cite{cui2022democratizing} introduce fine-grained late interaction between the two modalities; CyCLIP~\cite{goel2022cyclip} enforces geometric consistency through the usage of two additional objectives on top of the standard InfoNCE loss; SigLIP~\cite{zhai2023sigmoid} and SigLIP2~\cite{tschannen2025siglip} explore alternative contrastive objectives; SoftCLIP~\cite{gao2024softclip} relaxes the strict one-to-one alignment assumption by introducing soft cross-modal targets derived from intra-modal similarities; SILC~\cite{naeem2024silc} enhances representation quality via self-distillation; LaCLIP~\cite{fan2023improving} leverages language-centric augmentation and TULIP~\cite{Tang_2025_ICCV} enhances fine-grained visual understanding while preserving semantic alignment by combining generative data augmentation, intra-modal contrastive learning, and reconstruction-based regularization. In our work, we integrate multimodal and unimodal contrastive losses through a novel temperature modulation framework.

\myparagraph{Temperature in CL} has also received considerable attention, and recent work~\cite{wang2021understanding, qiu2024cool, kim2026temperature, li2023curriculum, wang2020contextual, zhang2021temperature, khaertdinov2022dynamic, kukleva2023temperature} has shown that the temperature parameter $\tau$ plays a crucial role in shaping the embedding space in both unimodal and multimodal contrastive learning. In the unimodal setting, \cite{wang2021understanding} showed that a low temperature concentrates the contrastive objective on hard negatives, while a high temperature distributes the objective more evenly across negatives, tolerating semantically similar samples and encouraging the formation of larger clusters~\cite{kukleva2023temperature}. Temperature therefore provides an implicit form of hardness weighting at the loss level, complementary to explicit hard negative mining, which instead changes the sampling distribution by favoring difficult examples~\cite{robinson2020contrastive}. A global $\tau$, however, applies the same weighting to all pairs and couples the treatment of positive and negative pairs, motivating approaches that adapt or remove the temperature parameter. In this direction, Temperature Schedules~\cite{kukleva2023temperature} replace a fixed $\tau$ with a cosine schedule that alternates between instance- and group-wise discrimination, while MM-TS~\cite{sheludzko2026mm} extends temperature scheduling to multimodal contrastive learning by dynamically adjusting the temperature during training and adapting its magnitude to the local data distribution. DySTreSS~\cite{manna2025dynamically} instead sets a pairwise temperature as a cosine function of similarity; MACL~\cite{huang2023model} makes temperature alignment-aware; Dynamic Temperature Scaling~\cite{khaertdinov2022dynamic} derives instance-level temperatures for negative pairs using a frozen auxiliary encoder; and Temperature-Free CL~\cite{kim2026temperature} removes the temperature hyperparameter altogether by replacing the usual scaling of similarity logits with a monotone log-odds mapping. Beyond its role in hardness weighting, temperature also influences the modality gap in multimodal contrastive learning, with \cite{yaras2024explaining} providing theoretical insights into this relationship. Building on these findings, we extend per-pair, similarity-driven temperature modulation to the multimodal setting and combine it with self-supervision to explicitly address the modality gap without introducing additional training stages.
\section{Method}
We introduce \textbf{TeMo}, a contrastive learning framework that assigns
similarity-conditioned temperatures to individual sample pairs and
progressively incorporates them into multimodal contrastive training.
We first introduce the notation and contrastive objectives underlying our
framework. We then present the proposed temperature modulation mechanism,
show how it is applied to both unimodal and multimodal pairs, and finally
combine them objectives through a progressive training schedule.

\myparagraph{Overview.}
TeMo augments standard multimodal contrastive learning with pair-specific
temperature modulation that is introduced progressively over the course of
training. At the beginning of training, learning is dominated by a
fixed-temperature symmetric InfoNCE objective, which establishes an initial
cross-modal alignment and organizes the global embedding space. As training
progresses, TeMo gradually increases the contribution of
temperature-modulated objectives defined over both cross-modal and
within-modality pairs.

\begin{figure*}[t]
    \centering
    \includegraphics[width=0.9\textwidth]{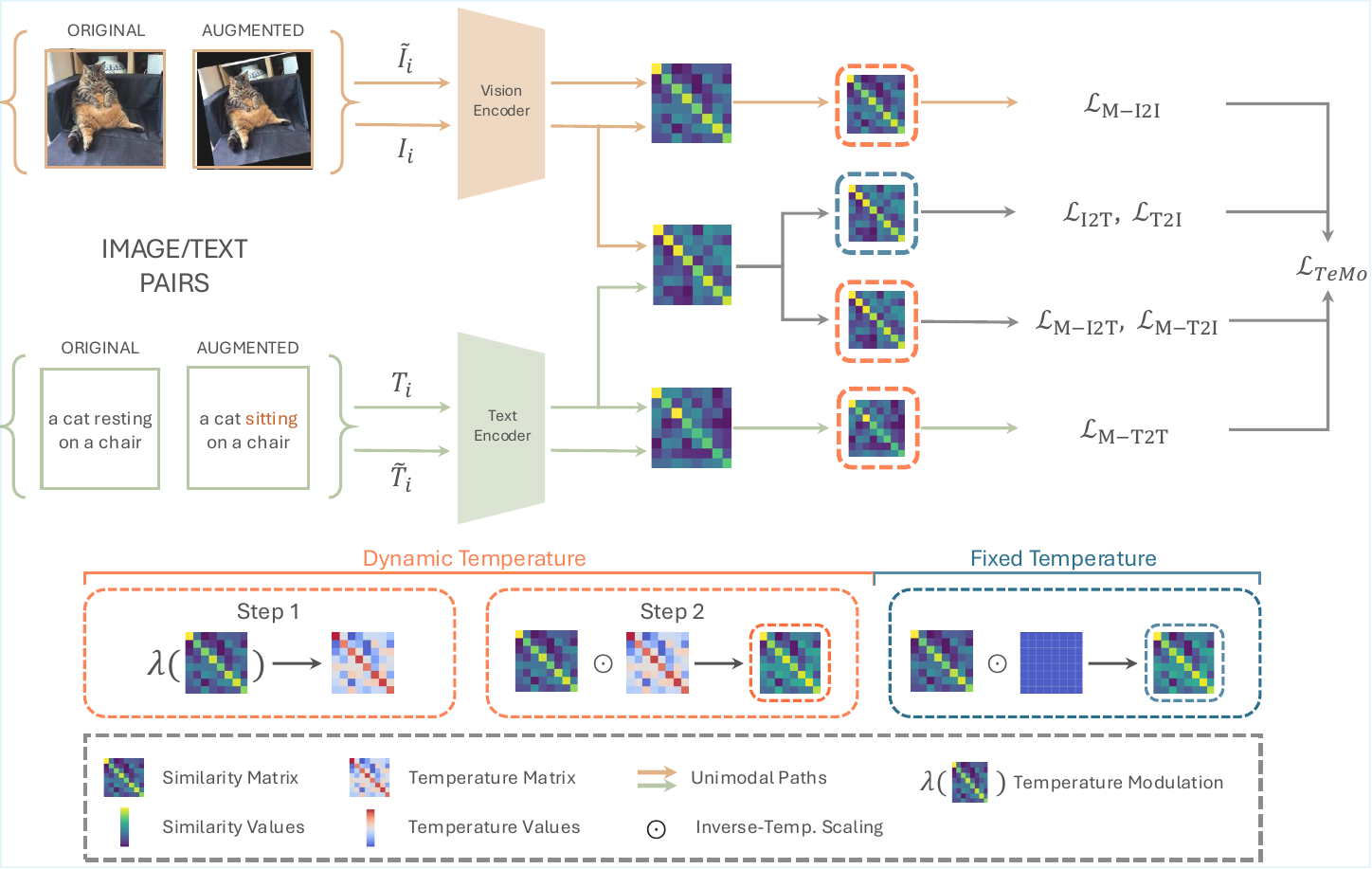}
    \caption{
        \textbf{High-level overview of TeMo.}
        A batch of image--text pairs and their augmentations is processed by a
        vision encoder and a text encoder to obtain embeddings
        $\mathbf{I}_i$, $\tilde{\mathbf{I}}_i$, $\mathbf{T}_i$, and
        $\tilde{\mathbf{T}}_i$. A fixed-temperature branch computes the standard
        cross-modal objective $\mathcal{L}_{\mathrm{MM}}$. In parallel, similarity
        matrices for image-to-text, text-to-image, image-to-image, and text-to-text
        pairs are passed through the temperature modulation mechanism, which maps
        similarities in $[0,1]$ to temperatures in
        $[\tau_{\min},\,\tau_{\min}+\tau_{\alpha}]$. The resulting temperatures
        are used to compute the modulated cross-modal loss
        $\mathcal{L}_{\mathrm{M\text{-}MM}}$ and the unimodal losses
        $\mathcal{L}_{\mathrm{M\text{-}I2I}}$ and
        $\mathcal{L}_{\mathrm{M\text{-}T2T}}$.
        Finally, a quadratic scheduler progressively shifts the training objective
        from the standard contrastive loss toward the temperature-modulated losses,
        yielding the final objective $\mathcal{L}_{\mathrm{TeMo}}$.
    }
    \label{fig:main_image}
\end{figure*}

Rather than applying a single temperature to all pairs in a minibatch, these
objectives assign a temperature according to the similarity of each
anchor--candidate pair. In particular, less similar pairs receive lower
temperatures, whereas more similar pairs receive higher temperatures,
sharpening or softening their corresponding contrastive logits. By introducing
this modulation progressively, TeMo first learns a stable global
representation structure before increasingly emphasizing finer pairwise
relationships within and across modalities. An overview of the framework is
shown in Figure~\ref{fig:main_image}.

\myparagraph{Notation.}
Let $\mathcal{D}={(I_n,T_n)}_{n=1}^{M}$ denote a multimodal dataset of $M$
paired image--text samples. For a minibatch of size $B$, we encode the images
and texts into the representations
$\mathbf{V}=(\mathbf{v}_1,\ldots,\mathbf{v}_B)$ and
$\mathbf{T}=(\mathbf{t}_1,\ldots,\mathbf{t}_B)$, respectively. For each image $I_i$,
we sample an augmented view $\tilde{I}_i\sim\mathcal{A}(I_i)$, and for each
text $T_i$, we generate an LLM-based paraphrase $\tilde{T}_i$. We denote the
representations of the augmented samples by
$\widetilde{\mathbf{V}}=(\tilde{\mathbf{v}}_1,\ldots,\tilde{\mathbf{v}}_B)$ and
$\widetilde{\mathbf{T}}=(\tilde{\mathbf{t}}_1,\ldots,\tilde{\mathbf{t}}_B)$. For any representation $\mathbf{x}$, we denote its $\ell_2$-normalized form by
$\bar{\mathbf{x}}=\mathbf{x}/\lVert\mathbf{x}\rVert_2$. The cosine similarity
between two representations $\mathbf{x}$ and $\mathbf{y}$ is
$c(\mathbf{x},\mathbf{y})=\langle\bar{\mathbf{x}},\bar{\mathbf{y}}\rangle$,
with $c(\mathbf{x},\mathbf{y})\in[-1,1]$.

\subsection{Preliminaries}

\myparagraph{Contrastive Loss.}
We adopt InfoNCE~\cite{oord2018representation} as the underlying contrastive
objective. Let $\mathbf{x}_i$ be an anchor, $\mathbf{Y}=\{\mathbf{y}_j\}_{j=1}^{B}$ a
candidate set, and $\mathbf{y}_i\in\mathbf{Y}$ its corresponding positive.
Let $c_{i,j}=c(\mathbf{x}_i,\mathbf{y}_j)$ denote the cosine similarity between
the anchor and candidate $j$, and let
$\boldsymbol{\tau}_i=(\tau_{i,1},\ldots,\tau_{i,B})$ denote the temperatures
associated with the $B$ anchor--candidate pairs. The InfoNCE loss for anchor
$\mathbf{x}_i$ is given by:
{\small
\begin{equation}
    \mathcal{I}
    \left(
        \mathbf{x}_i,\mathbf{Y};
        \boldsymbol{\tau}_i
    \right)
    =
    -\log
    \frac{
        \exp\left(c_{i,i}/\tau_{i,i}\right)
    }{
        \sum_{j=1}^{B}
        \exp\left(c_{i,j}/\tau_{i,j}\right)
    }.
    \label{eq:infonce}
\end{equation}
}

Standard InfoNCE applies the same fixed or learnable temperature $\tau_0$ to
every pair, such that
$\boldsymbol{\tau}_i=\tau_0\mathbf{1}_B$, where $\mathbf{1}_B$ is a vector of
ones of length $B$. This shared-temperature formulation serves as the basis of
standard contrastive training. TeMo generalizes it by allowing the
temperature to vary across individual anchor--candidate pairs.

\myparagraph{Unimodal Contrastive Loss.}
We first express the standard contrastive objective within a single modality.
Let $\mathbf{X}=\{\mathbf{x}_i\}_{i=1}^{B}$ denote a batch of embeddings from
one modality and
$\widetilde{\mathbf{X}}=\{\tilde{\mathbf{x}}_i\}_{i=1}^{B}$ its augmented
counterpart, where $\tilde{\mathbf{x}}_i$ forms the positive pair for
$\mathbf{x}_i$. The corresponding fixed-temperature unimodal objective is given by:
{\small
\begin{equation}
    \mathcal{L}_{\mathrm{X2X}}
    \left(
        \mathbf{X},
        \widetilde{\mathbf{X}};
        \tau_0
    \right)
    =
    \frac{1}{B}
    \sum_{i=1}^{B}
    \mathcal{I}
    \left(
        \mathbf{x}_i,
        \widetilde{\mathbf{X}};
        \tau_0\mathbf{1}_B
    \right),
    \label{eq:unimodal_infonce}
\end{equation}
}
where $\mathbf{X}$ represents either modality. In particular,
$(\mathbf{X},\widetilde{\mathbf{X}})
=(\mathbf{V},\widetilde{\mathbf{V}})$ defines the image-to-image objective,
while
$(\mathbf{X},\widetilde{\mathbf{X}})
=(\mathbf{T},\widetilde{\mathbf{T}})$ defines the corresponding text-to-text objective.

\myparagraph{Multimodal Contrastive Loss.}
Extending the contrastive formulation across modalities, we define the
fixed-temperature multimodal objective in both the image-to-text and
text-to-image directions:
{\small
\begin{equation}
    \begin{aligned}
        \mathcal{L}_{\mathrm{I2T}}
        \left(
            \mathbf{V},
            \mathbf{T};
            \tau_0
        \right)
        &=
        \frac{1}{B}
        \sum_{i=1}^{B}
        \mathcal{I}
        \left(
            \mathbf{v}_i,
            \mathbf{T};
            \tau_0\mathbf{1}_B
        \right),
        \\
        \mathcal{L}_{\mathrm{T2I}}
        \left(
            \mathbf{T},
            \mathbf{V};
            \tau_0
        \right)
        &=
        \frac{1}{B}
        \sum_{i=1}^{B}
        \mathcal{I}
        \left(
            \mathbf{t}_i,
            \mathbf{V};
            \tau_0\mathbf{1}_B
        \right).
    \end{aligned}
    \label{eq:directional_multimodal_infonce}
\end{equation}
}

Following standard multimodal contrastive learning, we average the two
directions to obtain the symmetric objective:
\begin{equation}
    \mathcal{L}_{\mathrm{MM}}
    =
    \frac{1}{2}
    \left(
        \mathcal{L}_{\mathrm{I2T}}
        +
        \mathcal{L}_{\mathrm{T2I}}
    \right).
    \label{eq:multimodal_infonce}
\end{equation}

The fixed-temperature objective supplies TeMo's initial training signal, but applies the same temperature to every pair. To refine the resulting representation, we next condition the temperature on the similarity of each pair.

\subsection{Temperature Modulation Framework}
\myparagraph{Temperature Modulation.}
The central component of TeMo is a similarity-dependent mapping that assigns
a separate temperature to each anchor--candidate pair. Let
$s_{i,j}=\tfrac{1}{2}\bigl(1+c(\mathbf{x}_i,\mathbf{y}_j)\bigr)\in[0,1]$ be the
cosine similarity rescaled to the unit interval. The temperature assigned to each pair
$(\mathbf{x}_i,\mathbf{y}_j)$ is given by:
\begin{equation}
    \tau_{i,j}^{\mathrm{X2Y}}(\mathbf{x}_i,\mathbf{y}_j)
    =
    \tau_{\min} + \tau_{\alpha}\sqrt{s_{i,j}},
    \label{eq:temperature_mapper}
\end{equation}
where $\tau_{\min}>0$ specifies the minimum temperature and
$\tau_{\alpha}\geq0$ controls the range of modulation, so that
$\tau_{i,j}^{\mathrm{X2Y}}\in[\tau_{\min},\tau_{\min}+\tau_{\alpha}]$. Because the mapping from similarity to temperature is monotonically increasing, less similar pairs are assigned lower temperatures, whereas more similar pairs receive
higher temperatures. Consequently, TeMo modifies the sharpness of the contrastive logits
on a pair-by-pair basis rather than applying the same scaling uniformly
throughout the minibatch.

Applying Equation \eqref{eq:temperature_mapper} to every pair formed by two embedding sets
$\mathbf{X}=\{\mathbf{x}_i\}_{i=1}^{B}$ and
$\mathbf{Y}=\{\mathbf{y}_j\}_{j=1}^{B}$ produces the temperature matrix:
{\small
\begin{equation}
    \mathcal{T}_{\mathrm{X2Y}}
    =
    \left[
        \tau_{i,j}^{\mathrm{X2Y}}
    \right]_{i,j=1}^{B}.
    \label{eq:temperature_matrix}
\end{equation}
}

The $i$-th row
$\boldsymbol{\tau}_i^{\mathrm{X2Y}}
=(\tau_{i,1}^{\mathrm{X2Y}},\ldots,\tau_{i,B}^{\mathrm{X2Y}})$ therefore
contains all temperatures used when $\mathbf{x}_i$ acts as the anchor.
This formulation can be applied directly to both within-modality and
cross-modal contrastive objectives.

\paragraph{Unimodal Temperature Modulation.}
We first apply the pair-specific temperatures within each modality. For a
batch $\mathbf{X}$ and its augmented counterpart $\widetilde{\mathbf{X}}$, we
form the temperature matrix
$\mathcal{T}_{\mathrm{X2X}}=\big[\tau_{i,j}^{\mathrm{X2X}}\big]_{i,j=1}^{B}$,
whose entries follow Eq.~\ref{eq:temperature_mapper} applied to the pair
$(\mathbf{x}_i,\tilde{\mathbf{x}}_j)$. Replacing the shared temperature in
Eq.~\ref{eq:unimodal_infonce} with the corresponding row of this matrix
yields the modulated unimodal objective:
{\small
\begin{equation}
    \mathcal{L}_{\mathrm{M\text{-}X2X}}
    \left(
        \mathbf{X}, \widetilde{\mathbf{X}}; \mathcal{T}_{\mathrm{X2X}}
    \right)
    =
    \frac{1}{B}
    \sum_{i=1}^{B}
    \mathcal{I}
    \left(
        \mathbf{x}_i, \widetilde{\mathbf{X}}; \boldsymbol{\tau}_i^{\mathrm{X2X}}
    \right).
    \label{eq:modulated_unimodal_loss}
\end{equation}
}
Applying it per modality gives the image and text objectives
$\mathcal{L}_{\mathrm{M\text{-}I2I}}=\mathcal{L}_{\mathrm{M\text{-}X2X}}(\mathbf{V},\widetilde{\mathbf{V}};\mathcal{T}_{\mathrm{I2I}})$
and
$\mathcal{L}_{\mathrm{M\text{-}T2T}}=\mathcal{L}_{\mathrm{M\text{-}X2X}}(\mathbf{T},\widetilde{\mathbf{T}};\mathcal{T}_{\mathrm{T2T}})$,
where $\mathcal{T}_{\mathrm{I2I}}$ and $\mathcal{T}_{\mathrm{T2T}}$ are formed
analogously from the image and text embeddings and their augmentations. These
objectives encourage each modality to preserve the semantics shared between a
sample and its augmentation, while letting the strength of each contrastive
interaction depend on its similarity.

\myparagraph{Multimodal Temperature Modulation.}
The same mechanism applies across modalities. The cross-modal temperature
matrix
$\mathcal{T}_{\mathrm{I2T}}=\big[\tau_{i,j}^{\mathrm{I2T}}\big]_{i,j=1}^{B}$
has entries following Eq.~\ref{eq:temperature_mapper} applied to the pair
$(\mathbf{v}_i,\mathbf{t}_j)$, and the reverse direction uses its transpose,
$\mathcal{T}_{\mathrm{T2I}}=\mathcal{T}_{\mathrm{I2T}}^{\top}$. Replacing the
shared temperature in the multimodal objective with the pair-specific
temperatures yields the directional losses:
{\small
\begin{equation}
    \begin{aligned}
        \mathcal{L}_{\mathrm{M\text{-}I2T}}
        \left(
            \mathbf{V}, \mathbf{T}; \mathcal{T}_{\mathrm{I2T}}
        \right)
        &=
        \frac{1}{B}
        \sum_{i=1}^{B}
        \mathcal{I}
        \left(
            \mathbf{v}_i, \mathbf{T}; \boldsymbol{\tau}_i^{\mathrm{I2T}}
        \right),
        \\
        \mathcal{L}_{\mathrm{M\text{-}T2I}}
        \left(
            \mathbf{T}, \mathbf{V}; \mathcal{T}_{\mathrm{T2I}}
        \right)
        &=
        \frac{1}{B}
        \sum_{i=1}^{B}
        \mathcal{I}
        \left(
            \mathbf{t}_i, \mathbf{V}; \boldsymbol{\tau}_i^{\mathrm{T2I}}
        \right),
    \end{aligned}
    \label{eq:crossmodal_modulated_loss_components}
\end{equation}
}
which we average, as in the fixed-temperature case, into the symmetric
modulated multimodal objective:
\begin{equation}
    \mathcal{L}_{\mathrm{M\text{-}MM}}
    =
    \frac{1}{2}
    \left(
        \mathcal{L}_{\mathrm{M\text{-}I2T}}
        +
        \mathcal{L}_{\mathrm{M\text{-}T2I}}
    \right).
    \label{eq:modulated_multimodal_loss}
\end{equation}
Together, the modulated multimodal and unimodal terms cover all pairwise
relationships in TeMo: image--text, image--image, and text--text. We collect
them into a single modulated objective,
\begin{equation}
    \mathcal{L}_{\mathrm{MOD}}
    =
    \mathcal{L}_{\mathrm{M\text{-}MM}}
    +
    \mathcal{L}_{\mathrm{M\text{-}I2I}}
    +
    \mathcal{L}_{\mathrm{M\text{-}T2T}}.
    \label{eq:modulated_total}
\end{equation}

\myparagraph{TeMo Objective.}
Applying all modulated objectives from the start of training would tie their
pair-specific weighting to similarities produced by encoders whose
representations are not yet reliable. TeMo therefore phases them in over the
course of training. Let $t\in[0,1]$ be the normalized training step; the
complete objective balances the standard multimodal loss with the modulated
one,
\begin{equation}
    \mathcal{L}_{\mathrm{TeMo}}(t)
    = \alpha(t)\,\mathcal{L}_{\mathrm{MM}}
    + \beta(t)\,\mathcal{L}_{\mathrm{MOD}},
    \label{eq:temo_loss}
\end{equation}
with weights $\alpha(t)=(1-t)^2$ and $\beta(t)=t^2$. At $t=0$ the objective
reduces to the standard multimodal loss $\mathcal{L}_{\mathrm{MM}}$, letting
the encoders establish an initial cross-modal representation before any
pair-specific modulation; as $t$ increases, the fixed-temperature term decays
quadratically while the modulated term grows, until at $t=1$ training is
driven entirely by $\mathcal{L}_{\mathrm{MOD}}$. This schedules a gradual
shift from global cross-modal organization toward finer pairwise refinement:
the modulated multimodal term continues to sharpen image--text alignment,
while the image--image and text--text terms structure each modality around the
semantics preserved under its augmentations. We summarize the full training procedure as pseudocode in Appendix~\ref{sec:temo_algorithm}.

\section{Experiments}

We evaluate TeMo on zero-shot cross-modal retrieval and image classification.
We first describe the experimental setup and compare TeMo with existing
contrastive and temperature-based approaches. We then ablate the individual
components of the proposed objective and conclude with an analysis of the
training dynamics and learned representation space.

\myparagraph{Datasets.}
We pretrain on Conceptual Captions using CC3M~\cite{sharma2018conceptual}
($\sim$2.9M image--text pairs) and its larger extension
CC12M~\cite{changpinyo2021conceptual} ($\sim$12M pairs). For zero-shot
retrieval, we evaluate on the Karpathy test splits~\cite{karpathy2015deep} of
MSCOCO and Flickr30k~\cite{young2014image}, following prior
work~\cite{yuan2022provable,goel2022cyclip}. For zero-shot classification, we evaluate on CIFAR-10, CIFAR-100~\cite{krizhevsky2009learning}, and ImageNet-1k~\cite{deng2009imagenet}, and on a broader 17-dataset suite from CLIP Benchmark~\cite{cherti_clip_bench} covering distribution shifts, fine-grained, and specialized domains.

\myparagraph{Evaluation.}
Zero-shot retrieval is evaluated using Recall@K for $K\in\{1,5,10\}$ in both
the image-to-text and text-to-image directions, with the complete results
reported in the supplementary. For zero-shot classification, we
report Top-K accuracy for $K\in\{1,3,5\}$ on CIFAR-10, CIFAR-100, and
ImageNet-1k, and Top-1 accuracy across the 17 CLIP Benchmark datasets.
Following the standard CLIP protocol, class names are used as textual prompts,
and each image is assigned to the class whose text representation has the
highest similarity to its image representation.

\myparagraph{Baselines.}
We compare TeMo against fixed, scheduled, and adaptive temperature methods, as
well as recent contrastive-learning variants. For the InfoNCE baseline, we use
a fixed temperature of $\tau=0.01$, following~\cite{qiu2024cool}. For the
temperature-scheduling baseline (TS$^{*}$), our multimodal adaptation of
Temperature Schedules~\cite{kukleva2023temperature}, we vary the temperature
within $\tau\in[0.01,0.05]$ across five training periods. For
SLIP~\cite{mu2021slip}, we use a learnable temperature initialized at $0.07$.
We further extend DySTreSS~\cite{manna2025dynamically} to the multimodal
setting and re-implement CWCL~\cite{srinivasa2023cwcl}, for which no public
implementation is available; both are trained using the same optimization
settings and backbone architectures as TeMo. For a fair comparison with
SoftCLIP~\cite{gao2024softclip}, we additionally train both the InfoNCE baseline and TeMo from
scratch on CC3M. For MM-TS~\cite{sheludzko2026mm}, we use the results reported in the original paper.

\myparagraph{Implementation Details.}
For experiments on CC3M, we initialize the vision and text encoders
independently rather than from a jointly pretrained CLIP model,
following~\cite{qiu2024cool}. We use either an ImageNet-1k-pretrained
ResNet-50~\cite{he2016deep} or ViT-B/16~\cite{dosovitskiy2020image} as the
vision encoder and a pretrained DistilBERT~\cite{sanh2019distilbert} as the
text encoder. For CC12M, we build upon the official
SLIP~\cite{mu2021slip} implementation and train the encoders from scratch.
For text augmentation, we pre-generate five paraphrases for each caption
using a PEGASUS~\cite{zhang2020pegasus} model fine-tuned for paraphrasing and
uniformly sample one variant at each training step. For TeMo, we set
$\tau_{\min}=0.01$ and $\tau_{\alpha}=0.04$ in
Eq.~\ref{eq:temperature_mapper}, resulting in temperatures within
$[0.01,0.05]$. We keep $\tau_{\min}$ fixed and treat only
$\tau_{\alpha}$ as a tunable hyperparameter. We use a batch size of 2048 for
CC3M and 4096 for CC12M.

\begin{table*}[t]
    \centering
    \small
    \setlength{\tabcolsep}{6pt}
    \renewcommand{\arraystretch}{1}
    \begin{tabular}{ll cc cc cc cc}
        \toprule
        \multirow{2}{*}{\textbf{Backbone}} &
        \multirow{2}{*}{\textbf{Method}}
        & \multicolumn{2}{c}{\textbf{MSCOCO}}
        & \multicolumn{2}{c}{\textbf{Flickr30k}}
        & \multicolumn{2}{c}{\textbf{CIFAR-10}}
        & \multicolumn{2}{c}{\textbf{CIFAR-100}} \\
        \cmidrule(lr){3-4}\cmidrule(lr){5-6}
        \cmidrule(lr){7-8}\cmidrule(lr){9-10}
        & & IR@1 & TR@1 & IR@1 & TR@1
        & Top-1 & Top-3 & Top-1 & Top-3 \\
        \midrule
        \multirow{6}{*}{RN50~\cite{he2016deep}}
            & InfoNCE
            & 21.64 & \underline{28.60} & 42.12 & 53.60
            & 53.78 & 81.76 & 25.08 & 42.09 \\
            & TS$^{*}$~\cite{kukleva2023temperature}
            & \underline{22.01} & 28.20 & \underline{42.90} & 53.30
            & 51.61 & 82.03 & 28.53 & 45.21 \\
            & DySTreSS$^{*}$~\cite{manna2025dynamically}
            & 19.38 & 26.70 & 36.70 & 49.30
            & 56.95 & \underline{85.49} & \underline{33.29}
            & \underline{52.55} \\
            & CWCL$^{*}$~\cite{srinivasa2023cwcl}
            & 14.83 & 23.50 & 28.70 & 41.40
            & \underline{57.92} & 80.84 & 28.46 & 46.72 \\
            & MM-TS~\cite{sheludzko2026mm}
            & 21.20 & 28.40 & 41.50 & \underline{54.30}
            & -- & -- & -- & -- \\
            \rowcolor{oursbg}
            & TeMo (ours)
            & \textbf{23.28} & \textbf{30.36}
            & \textbf{44.16} & \textbf{56.00}
            & \textbf{64.04} & \textbf{89.85}
            & \textbf{37.52} & \textbf{57.48} \\
        \midrule
        \multirow{5}{*}{ViT-B/16~\cite{dosovitskiy2020image}}
            & InfoNCE
            & \underline{21.88} & \textbf{28.98}
            & \underline{42.54} & \underline{54.10}
            & 70.55 & 90.21 & 44.53 & 61.41 \\
            & TS$^{*}$~\cite{kukleva2023temperature}
            & 21.45 & 27.32 & 41.02 & 49.90
            & 72.02 & 91.65 & 47.13 & 65.45 \\
            & DySTreSS$^{*}$~\cite{manna2025dynamically}
            & 20.67 & 26.26 & 39.36 & 49.70
            & 73.41 & 92.15 & 47.02 & 65.64 \\
            & CWCL$^{*}$~\cite{srinivasa2023cwcl}
            & 18.84 & 28.44 & 37.90 & 48.10
            & \textbf{85.85} & \textbf{95.18}
            & \textbf{62.02} & \textbf{78.63} \\
            \rowcolor{oursbg}
            & TeMo (ours)
            & \textbf{22.87} & \underline{28.80}
            & \textbf{45.28} & \textbf{55.60}
            & \underline{81.54} & \underline{94.47}
            & \underline{53.62} & \underline{70.39} \\
        \bottomrule
    \end{tabular}
    \caption{\textbf{Zero-shot cross-modal retrieval and classification.}
    Models are trained on CC3M; higher is better for all metrics. Retrieval on
    MSCOCO and Flickr30k is reported using R@1 (\%), with IR and TR denoting
    Text$\rightarrow$Image retrieval and Image$\rightarrow$Text retrieval,
    respectively. Zero-shot classification on CIFAR-10/100 reports Top-1 and
    Top-3 accuracy (\%). \textbf{Best} results are shown in bold and
    \underline{second-best} results are underlined; our method is highlighted.
    TS$^{*}$, DySTreSS$^{*}$, and CWCL$^{*}$ denote our multimodal
    adaptations/implementations of~\cite{kukleva2023temperature},
    \cite{manna2025dynamically}, and~\cite{srinivasa2023cwcl}, respectively.}
    \label{tab:sota_retrieval_and_zs_classification}
\end{table*}

\subsection{Comparison with State of the Art}

Table~\ref{tab:sota_retrieval_and_zs_classification} compares TeMo with prior
methods on zero-shot retrieval and classification across both backbone
architectures. TeMo achieves consistently strong retrieval performance across
datasets and architectures. With ResNet-50, it improves over InfoNCE by
+1.64/+1.76\% on MSCOCO and +2.04/+2.40\% on Flickr30k in IR@1/TR@1,
respectively. TeMo also outperforms all temperature-based baselines evaluated
in our study: TS~\cite{kukleva2023temperature},
DySTreSS~\cite{manna2025dynamically}, CWCL~\cite{srinivasa2023cwcl}, and
MM-TS~\cite{sheludzko2026mm}. Relative to MM-TS, the most closely related
multimodal temperature-scheduling approach to ours, TeMo improves IR@1/TR@1 by
+2.08/+1.96\% on MSCOCO and +2.66/+1.70\% on Flickr30k. These improvements
also transfer across architectures: while TS degrades when moving from
ResNet-50 to ViT-B/16, TeMo maintains consistently strong retrieval
performance across both backbones.

A similar trend is observed for zero-shot classification. Relative to
InfoNCE, TeMo improves Top-1 accuracy by +10.26/+12.44\% on
CIFAR-10/CIFAR-100 with ResNet-50 and by +10.99/+9.09\% with ViT-B/16.
Although CWCL attains higher classification accuracy with ViT-B/16, this is
accompanied by substantially lower retrieval performance; for instance, its
Flickr30k IR@1 is 7.38\% lower than that of TeMo. TeMo therefore provides a
more balanced performance across both tasks, combining strong retrieval
results with competitive zero-shot classification accuracy.

The benefits of TeMo persist when extending the evaluation to ImageNet-1k and
increasing the pretraining scale. On ImageNet-1k, TeMo achieves 34.79\%
Top-1 and 48.32\% Top-3 accuracy, outperforming both InfoNCE and TS
(Table~\ref{tab:cc12m_cc3m_zs}). We further compare against SoftCLIP using the
improvement over each method's corresponding InfoNCE baseline, since
differences in experimental setup lead to different absolute baseline
accuracies. SoftCLIP reports an improvement of +2.00 percentage points,
whereas TeMo trained from scratch on CC3M achieves +2.23 points
(Table~\ref{tab:softclip_gain}). Scaling pretraining from CC3M to CC12M
further increases TeMo's performance to 41.76\% Top-1 and 63.05\% Top-3
(Table~\ref{tab:cc12m_cc3m_zs}), surpassing InfoNCE, TS, and SLIP. These
results indicate that the improvements introduced by TeMo persist when
scaling to a larger pretraining dataset.

\begin{table}[t]
    \centering
    \small
    \setlength{\tabcolsep}{5pt}
    \renewcommand{\arraystretch}{1.0}
    \begin{tabular}{l cc cc}
        \toprule
        \multirow{2}{*}{\textbf{Method}}
        & \multicolumn{2}{c}{\textbf{CC3M}}
        & \multicolumn{2}{c}{\textbf{CC12M}} \\
        \cmidrule(lr){2-3}\cmidrule(lr){4-5}
        & Top-1 & Top-3 & Top-1 & Top-3 \\
        \midrule
        InfoNCE & 28.00 & 40.23 & 35.48 & 54.80 \\
        TS$^{*}$~\cite{kukleva2023temperature}
        & \underline{29.01} & \underline{42.17} & 38.90 & 59.02 \\
        SLIP~\cite{mu2021slip} & -- & -- & \underline{39.77} & \underline{59.42} \\
        \rowcolor{oursbg}
        TeMo (ours)
        & \textbf{34.79} & \textbf{48.32}
        & \textbf{41.76} & \textbf{63.05} \\
        \bottomrule
    \end{tabular}
    \caption{\textbf{Zero-shot classification on ImageNet-1k.}
    Top-1 and Top-3 accuracy (\%) of ViT-B/16 models pretrained on CC3M and
    CC12M. TS$^{*}$ denotes our multimodal adaptation of
    Temperature Schedules~\cite{kukleva2023temperature}.}
    \label{tab:cc12m_cc3m_zs}
\end{table}

\begin{table}[t]
    \centering
    \small
    \setlength{\tabcolsep}{6pt}
    \renewcommand{\arraystretch}{1.0}
    \begin{tabular}{l c c}
        \toprule
        \textbf{Method} & \textbf{Top-1}
        & \textbf{$\Delta$ vs.\ InfoNCE} \\
        \midrule
        InfoNCE$^{\dagger}$ & 16.90 & -- \\
        SoftCLIP$^{\dagger}$ & 18.90 & \underline{+2.00} \\
        \midrule
        InfoNCE & 14.39 & -- \\
        \rowcolor{oursbg}
        TeMo (ours) & 16.62 & \textbf{+2.23} \\
        \bottomrule
    \end{tabular}
    \caption{\textbf{Absolute-gain comparison with SoftCLIP on ImageNet-1k.}
    Top-1 accuracy (\%) for ViT-B/16 models trained from scratch on CC3M.
    Since absolute baselines differ across experimental setups, we compare
    each method's improvement over its corresponding InfoNCE baseline
    ($\Delta$). $^{\dagger}$Results are taken from the SoftCLIP paper.}
    \label{tab:softclip_gain}
\end{table}

\subsection{Ablations}

Table~\ref{tab:loss_components_ablation} ablates the main components of the
TeMo objective (Eq.~\ref{eq:temo_loss}), including multimodal temperature
modulation, unimodal supervision, and progressive scheduling. Row~a) denotes
the standard InfoNCE baseline, while row~g) corresponds to the complete TeMo
objective.

\begin{table}[t]
    \centering
    \small
    \setlength{\tabcolsep}{4pt}
    \renewcommand{\arraystretch}{1.0}
    \resizebox{\columnwidth}{!}{%
        \begin{tabular}{c cccc cc cc}
            \toprule
            & \multirow{2}{*}{\textbf{Base}}
            & \multirow{2}{*}{\textbf{Mod.}}
            & \multirow{2}{*}{\textbf{U.L.}}
            & \multirow{2}{*}{\textbf{Sch.}}
            & \multicolumn{2}{c}{\textbf{MSCOCO}}
            & \multicolumn{2}{c}{\textbf{Flickr30k}} \\
            \cmidrule(lr){6-7}\cmidrule(lr){8-9}
            & & & & & IR@1 & TR@1 & IR@1 & TR@1 \\
            \midrule
            a) & \checkmark & -- & -- & --
            & 21.64 & 28.60 & 42.12 & 53.60 \\
            b) & -- & \checkmark & -- & --
            & 18.32 & 25.36 & 36.16 & 45.00 \\
            c) & \checkmark & -- & \checkmark$^{\dagger}$ & --
            & 20.65 & 25.44 & 39.16 & 50.10 \\
            d) & \checkmark & \checkmark & -- & --
            & 20.09 & 27.54 & 40.12 & 52.20 \\
            e) & \checkmark & \checkmark & -- & \checkmark
            & 21.74 & 28.58 & \underline{44.02} & \underline{54.90} \\
            f) & \checkmark & \checkmark & \checkmark & --
            & \underline{21.90} & \underline{29.38} & 42.84 & 54.80 \\
            \rowcolor{oursbg}
            g) & \checkmark & \checkmark & \checkmark & \checkmark
            & \textbf{23.28} & \textbf{30.36}
            & \textbf{44.16} & \textbf{56.00} \\
            \bottomrule
        \end{tabular}
    }
    \caption{\textbf{Loss-component ablation.}
    Retrieval performance (\%) under different combinations
    of training objectives. Base: standard InfoNCE
    loss $\mathcal{L}_{\mathrm{MM}}$; Mod.: modulated multimodal loss
    $\mathcal{L}_{\mathrm{M\text{-}MM}}$; U.L.: additional unimodal losses;
    Sch.: progressive scheduler blending $\mathcal{L}_{\mathrm{MM}}$ with the
    modulated objectives. $^{\dagger}$Unimodal objectives are applied without
    temperature modulation.}
    \label{tab:loss_components_ablation}
\end{table}

\myparagraph{Multimodal Modulation.}
We first isolate the effect of multimodal temperature modulation by training
with $\mathcal{L}_{\mathrm{M\text{-}MM}}$ alone in row~b). Compared with the
standard InfoNCE objective in row~a), performance decreases across all
retrieval metrics, suggesting that applying pair-specific temperature
modulation from the beginning of training can interfere with the formation of
the initial cross-modal structure. We hypothesize that this behavior is
related to hard negatives. Since highly similar pairs are assigned higher
temperatures, the contrastive penalty associated with such negatives is
softened, which may limit their separation from the anchor before a meaningful
representation space has been established.

\myparagraph{Impact of Standard Unimodal Losses.}
To disentangle the contribution of unimodal supervision from that of
temperature modulation, row~c) augments the InfoNCE baseline with standard,
non-modulated unimodal objectives. This configuration performs below the
baseline in row~a) across all retrieval metrics, indicating that unimodal
supervision alone does not improve cross-modal alignment in our setting.
Prior work typically introduces unimodal supervision for a single modality,
such as language rewrites in LaCLIP~\cite{fan2023improving} or image
self-supervision in SLIP~\cite{mu2021slip}. Jointly optimizing separate
objectives for both modalities may instead alter the optimization dynamics of
the shared embedding space and interfere with cross-modal alignment. As shown
by the full TeMo objective in row~g), these unimodal objectives become
beneficial when combined with pair-specific temperature modulation and
progressive scheduling.

\myparagraph{Multimodal Modulation and Progressive Scheduling.}
We next examine how the standard multimodal objective
$\mathcal{L}_{\mathrm{MM}}$ interacts with its temperature-modulated
counterpart $\mathcal{L}_{\mathrm{M\text{-}MM}}$. Combining the two objectives
with fixed weights in row~d) does not improve upon the InfoNCE baseline in
row~a). In contrast, introducing the same modulated objective progressively
through the proposed scheduler in row~e) substantially improves over the
fixed-weight formulation, yielding gains of +1.65/+1.04\% on MSCOCO and
+3.90/+2.70\% on Flickr30k in IR@1/TR@1, respectively. The same trend is
observed when unimodal objectives are included, with row~g) consistently
improving over its unscheduled counterpart in row~f).

We attribute this behavior to the complementary roles of the objectives over
the course of training. During the early stages, the scheduler places greater
emphasis on $\mathcal{L}_{\mathrm{MM}}$, whose low fixed temperature provides
a strong contrastive signal for separating anchors from negatives. As the
representation space becomes more structured, the contribution of the
temperature-modulated objectives increases, allowing pair-specific
temperatures to refine the learned semantic relationships. The improvements
from row~d) to row~e) and from row~f) to row~g) therefore demonstrate the
importance of introducing temperature modulation progressively rather than
assigning it a fixed contribution throughout training.

\myparagraph{Unimodal Modulation.}
Finally, we assess the contribution of the temperature-modulated unimodal
objectives by comparing the scheduled multimodal variant in row~e) with the
complete TeMo objective in row~g). Adding
$\mathcal{L}_{\mathrm{M\text{-}I2I}}$ and
$\mathcal{L}_{\mathrm{M\text{-}T2T}}$ further improves performance across all
retrieval metrics, with gains of +1.54/+1.78\% on MSCOCO and +0.14/+1.10\%
on Flickr30k in IR@1/TR@1, respectively. Together with the degradation
observed when standard unimodal objectives are used without modulation in
row~c), these results indicate that the benefit of unimodal supervision
depends on how it is integrated into the contrastive objective. The complete
TeMo formulation in row~g) achieves the strongest overall retrieval
performance, improving over the InfoNCE baseline in row~a) by +1.64/+1.76\%
on MSCOCO and +2.04/+2.40\% on Flickr30k.

The same trend extends beyond retrieval. Across the 17 CLIP Benchmark
datasets, row~g) achieves a macro-average Top-1 accuracy of 26.98\%, compared
with 22.54\% for the InfoNCE baseline in row~a), corresponding to an
improvement of +4.44 percentage points. The per-dataset results are reported in Table~\ref{tab:loss_components_zsh_summary}.

\subsection{Representation and Training Dynamics}

\begin{figure}[t]
    \centering
    \includegraphics[width=\columnwidth]{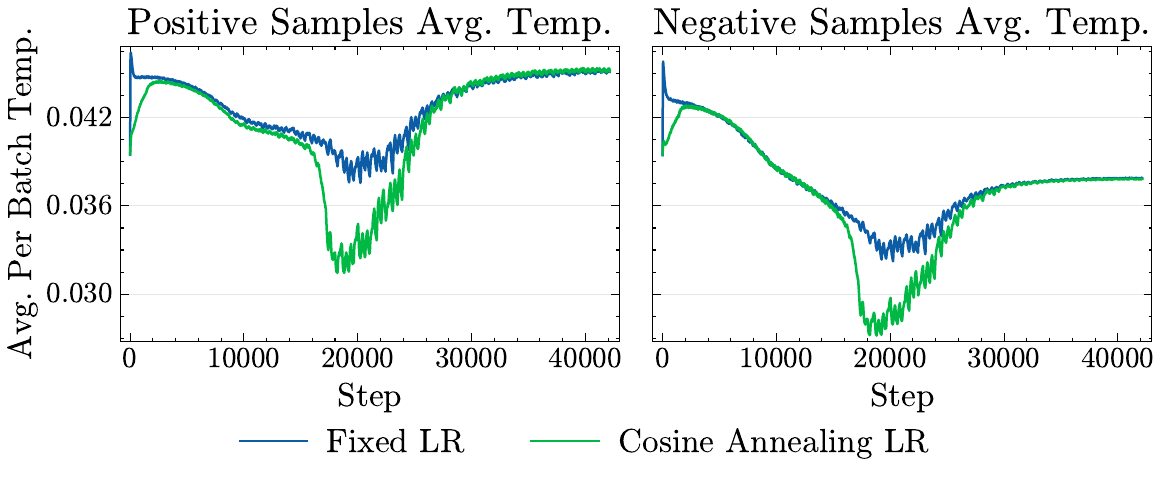}
    \caption{\textbf{Convergence of batch-wise temperatures.}
    Average temperatures assigned to positive and negative pairs converge
    toward stable values during training under both cosine-annealed and
    fixed learning-rate schedules.}
    \label{fig:temperature_convergence}
\end{figure}

\myparagraph{Temperature Convergence.}
To better understand the behavior of the proposed modulation, we track the
average temperature assigned to positive and negative pairs within each
minibatch throughout training, as shown in
Figure~\ref{fig:temperature_convergence}. Both exhibit an initial decrease,
followed by gradual stabilization as the contribution of the modulated
objectives increases. By the end of training, positive pairs consistently
receive higher temperatures than negative pairs, reflecting the higher
similarity of aligned image--text pairs under the proposed mapping. To verify
that this behavior is not induced by the cosine-annealing learning-rate
schedule, we repeat the experiment using a fixed learning rate and observe the
same convergence pattern. This suggests that the stabilization of the
temperatures arises from the learned representation structure rather than from
the learning-rate schedule.

\begin{figure}[t]
    \centering
    \includegraphics[width=\columnwidth]{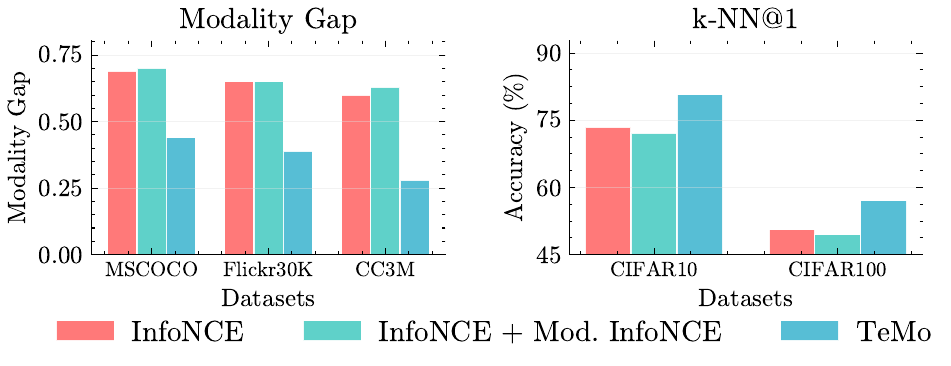}
    \caption{\textbf{Modality gap and $k$-NN@1 evaluation.}
    Left: modality gap across evaluation datasets.
    Right: $k$-NN@1 accuracy of the learned visual representations on
    CIFAR-10 and CIFAR-100.}
    \label{fig:modality_gap_and_knn}
\end{figure}

\myparagraph{Modality Gap and $k$-NN Evaluation.}
We further analyze the influence of the unimodal objectives on both
cross-modal alignment and the structure of the individual representation
spaces. Following~\cite{liang2022mind}, we measure the modality gap and
evaluate the visual representations using $k$-NN classification
(Figure~\ref{fig:modality_gap_and_knn}). We compare the standard InfoNCE
baseline with variants incorporating multimodal modulation and unimodal
objectives. Introducing the unimodal objectives improves $k$-NN accuracy,
indicating a stronger visual representation space, whereas multimodal
temperature modulation alone has only a limited effect on visual
representation quality. The complete TeMo objective additionally yields the
smallest modality gap among the evaluated variants, suggesting improved
cross-modal alignment. These observations are consistent with the component
ablation in Table~\ref{tab:loss_components_ablation}.

\begin{figure}[t]
    \centering
    \includegraphics[width=0.75\columnwidth]
    {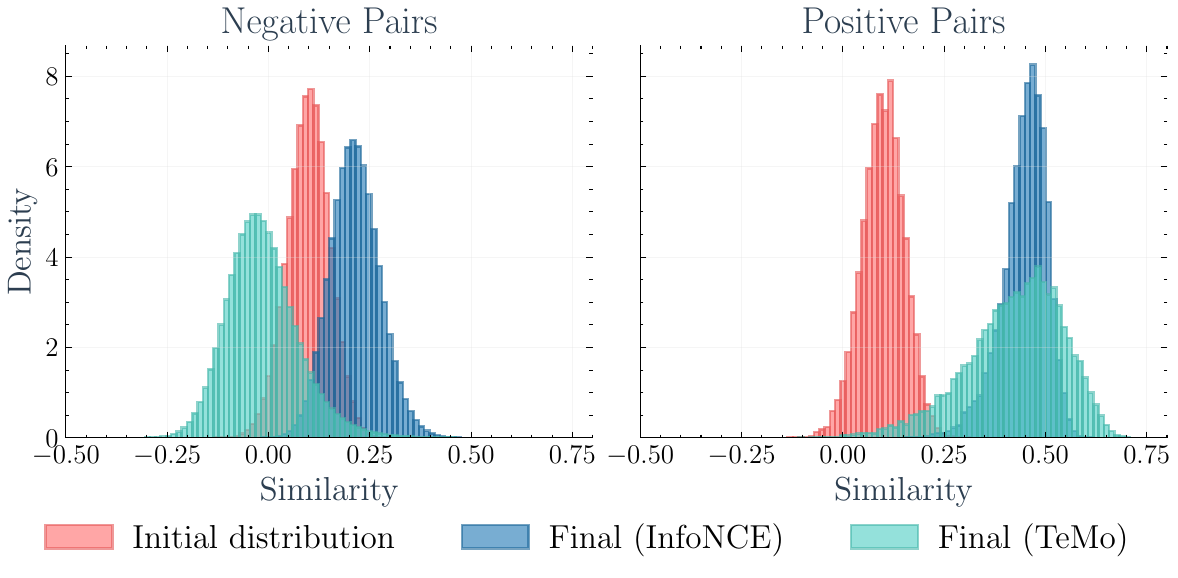}
    \caption{\textbf{Normalized distributions of image--text similarity on the
    CC3M validation set for negative (left) and positive (right) pairs.}
    Red denotes the distributions before fine-tuning, blue after InfoNCE
    training, and green after training with TeMo.}
    \label{fig:similarity_distribution}
\end{figure}

\myparagraph{Distribution of Image--Text Similarity.}
Figure~\ref{fig:similarity_distribution} examines the similarity distributions
of positive and negative image--text pairs on the CC3M validation set. Before
fine-tuning, the two distributions overlap substantially, indicating limited
separation between aligned and non-aligned pairs. Training with InfoNCE shifts
the positive distribution toward higher similarities; however, the negative
distribution also shifts in the same direction, leaving a considerable region
of overlap between hard negatives and positive pairs. In contrast, TeMo shifts
positive pairs toward higher similarities while moving negative pairs toward
lower similarities, thereby increasing the separation between the two
distributions. The resulting representation space exhibits a clearer margin
between positive and negative pairs. We provide a more detailed analysis of
robustness to noisy positive pairs in
Appendix~\ref{sec:robustness_agains_fn_sec}. Overall, these results indicate
that TeMo produces a more discriminative cross-modal representation space,
consistent with the improvements observed in retrieval and zero-shot
classification.
\section{Conclusions}
In this work, we proposed {TeMo}, a \underline{Te}mperature \underline{Mo}dulation method for multimodal Contrastive Learning, which aims to improve the standard multimodal contrastive learning objective. 
First, we introduced an adaptive per-pair temperature modulation method, where the temperature for each training sample is adapted based on its local neighborhood in the embedding space. This allows the model to adjust the pushing force of each contrastive pair more effectively, adapting it to the respective neighborhood. Next, we extended this idea to the unimodal domain by integrating temperature modulation into the unimodal contrastive losses. These components are then combined into a unified modulated contrastive loss formulation, which further reinforces representation learning within each modality. Finally, we blended the modulated contrastive loss with the original InfoNCE objective with the progressive scheduling. This combination enables learning of instance-level details using a lower temperature, while our adaptive modulation framework gradually enables coarser semantic grouping.
Our results demonstrate consistent improvements across retrieval on MSCOCO and Flickr30K and zero-shot classification on a diverse set of datasets, including CIFAR10, CIFAR100, and ImageNet-1k, as well as across evaluation metrics, highlighting the effectiveness of our method.

{
    \small
    \bibliographystyle{ieeenat_fullname}
    \bibliography{main}
}

\clearpage
\maketitlesupplementary

In the supplementary material, we first present the full TeMo training
algorithm (Sec.~\ref{sec:temo_algorithm}). We then provide extended versions of
the main-paper tables, covering state-of-the-art comparisons on retrieval and
classification, the loss-component ablations on both tasks, and $k$-NN
evaluation of the visual representations. We further include additional
analyses: a connection between TeMo and the alignment and uniformity framework,
its effect on the smoothness of the objective, the role of the similarity
source (model-derived versus expert-derived), the robustness of TeMo to false
positives, and a discussion of its limitations.

\section{TeMo Algorithm}
\label{sec:temo_algorithm}
\begin{algorithm}[h]
  \caption{TeMo algorithm}
  \label{alg:temo}
  \begin{algorithmic}[1]
    \REQUIRE {
      $\mathbf{I}$ - batch of images;
      $\mathbf{T}$ - batch of texts;
      $\mathbf{\tilde{I}}$ - batch of augmented images;
      $\mathbf{\tilde{T}}$ - batch of augmented texts;
      $\tau_0$ - fixed baseline temperature;
      $t$ - normalized training step ($t \in [0, 1]$);
      $\tau_{\min}$ - minimum temperature;
      $\tau_{\alpha}$ - temperature scaling factor.
    }
    \FOR{$(\mathbf{I}, \mathbf{T}, \mathbf{\tilde{I}}, \mathbf{\tilde{T}})$ in loader}
      \STATE \textbf{\# Standard multimodal contrastive losses}
      \STATE $\mathcal{L}_{\text{I2T}} \leftarrow \mathcal{I}(\mathbf{I}, \mathbf{T}, \tau_0)$
      \STATE $\mathcal{L}_{\text{T2I}} \leftarrow \mathcal{I}(\mathbf{T}, \mathbf{I}, \tau_0)$
      \STATE $\mathcal{L}_{\text{MM}} \leftarrow \frac{1}{2}(\mathcal{L}_{\text{I2T}} + \mathcal{L}_{\text{T2I}})$
      \STATE
      \STATE \textbf{\# Modulated multimodal losses}
      \STATE $\mathcal{T}_{\text{I2T}} \leftarrow \tau_{\min} + \tau_{\alpha}\sqrt{s(\mathbf{I}, \mathbf{T})}$
      \STATE $\mathcal{T}_{\text{T2I}} \leftarrow \mathcal{T}_{\text{I2T}}^{\top}$
      \STATE $\mathcal{L}_{\text{M-I2T}} \leftarrow \mathcal{I}(\mathbf{I}, \mathbf{T}, \mathcal{T}_{\text{I2T}})$
      \STATE $\mathcal{L}_{\text{M-T2I}} \leftarrow \mathcal{I}(\mathbf{T}, \mathbf{I}, \mathcal{T}_{\text{T2I}})$
      \STATE $\mathcal{L}_{\text{M-MM}} \leftarrow \frac{1}{2}(\mathcal{L}_{\text{M-I2T}} + \mathcal{L}_{\text{M-T2I}})$
      \STATE
      \STATE \textbf{\# Modulated unimodal losses}
      \STATE $\mathcal{T}_{\text{I2I}} \leftarrow \tau_{\min} + \tau_{\alpha}\sqrt{s(\mathbf{I}, \mathbf{\tilde{I}})}$
      \STATE $\mathcal{T}_{\text{T2T}} \leftarrow \tau_{\min} + \tau_{\alpha}\sqrt{s(\mathbf{T}, \mathbf{\tilde{T}})}$
      \STATE $\mathcal{L}_{\text{M-I2I}} \leftarrow \mathcal{I}(\mathbf{I}, \mathbf{\tilde{I}}, \mathcal{T}_{\text{I2I}})$
      \STATE $\mathcal{L}_{\text{M-T2T}} \leftarrow \mathcal{I}(\mathbf{T}, \mathbf{\tilde{T}}, \mathcal{T}_{\text{T2T}})$
      \STATE
      \STATE \textbf{\# Quadratic scheduler}
      \STATE $\alpha \leftarrow (1 - t)^2, \quad \beta \leftarrow t^2$
      \STATE
      \STATE \textbf{\# Final TeMo loss}
      \STATE $\mathcal{L}_{\text{TeMo}} \leftarrow \alpha \cdot \mathcal{L}_{\text{MM}} + \beta \cdot (\mathcal{L}_{\text{M-MM}} + \mathcal{L}_{\text{M-I2I}} + \mathcal{L}_{\text{M-T2T}})$
    \ENDFOR
  \end{algorithmic}
\end{algorithm}

Here $s(\mathbf{A},\mathbf{B})$ denotes the matrix of pairwise normalized similarities between two sets, and all operations are applied elementwise.

\section{Comparison to State-Of-The-Art}
Tables~\ref{tab:sota_retrieval_extended} and ~\ref{tab:sota_zshot_extended} show a more detailed comparison with state-of-the-art models on zero-shot retrieval and zero-shot classification tasks. In addition to the metrics reported in the main paper, we also report R@5, R@10, and R-Mean (the average of the image-to-text and text-to-image) on MSCOCO and Flickr30k, as well as Top-3 and Top-5 accuracy on CIFAR-10, CIFAR-100, and ImageNet-1k. TeMo consistently delivers the strongest performance across every metric.

\begin{table*}[ht]
    \centering
    \small
    \setlength{\tabcolsep}{4pt}
    \renewcommand{\arraystretch}{1.1}
    \resizebox{\textwidth}{!}{%
        \begin{tabular}{ll ccc ccc c ccc ccc c}
            \toprule
            \multirow{2}{*}{\textbf{Backbone}} & \multirow{2}{*}{\textbf{Method}}
            & \multicolumn{3}{c}{\textbf{MSCOCO IR}} & \multicolumn{3}{c}{\textbf{MSCOCO TR}}
            & \multirow{2}{*}{\textbf{R-Mean}}
            & \multicolumn{3}{c}{\textbf{Flickr30k IR}} & \multicolumn{3}{c}{\textbf{Flickr30k TR}}
            & \multirow{2}{*}{\textbf{R-Mean}} \\
            \cmidrule(lr){3-5}\cmidrule(lr){6-8}\cmidrule(lr){10-12}\cmidrule(lr){13-15}
            & & R@1 & R@5 & R@10 & R@1 & R@5 & R@10 & & R@1 & R@5 & R@10 & R@1 & R@5 & R@10 & \\
            \midrule
            \multirow{8}{*}{RN50}
                & InfoNCE & 21.64 & \underline{45.77} & 57.47 & \underline{28.60} & 53.90 & 66.56 & 45.66 & 42.12 & 68.84 & 78.76 & 53.60 & \underline{81.30} & 88.90 & 68.92 \\
                & TS$^{*}$~\cite{kukleva2023temperature} & \underline{22.01} & 45.51 & \underline{57.50} & 28.20 & \underline{54.62} & \underline{66.80} & \underline{45.77} & \underline{42.90} & \underline{69.92} & \underline{79.48} & 53.30 & 81.10 & \underline{89.10} & \underline{69.30} \\
                & DySTreSS$^{*}$~\cite{manna2025dynamically} & 19.38 & 42.34 & 54.27 & 26.70 & 52.24 & 64.02 & 43.16 & 36.70 & 64.98 & 75.66 & 49.30 & 77.80 & 87.10 & 65.26 \\
                & DySTreSS Shifted$^{*}$~\cite{manna2025dynamically} & 19.31 & 42.53 & 54.33 & 25.12 & 50.38 & 62.52 & 42.37 & 37.22 & 65.22 & 75.88 & 45.70 & 76.00 & 85.00 & 64.17 \\
                & CWCL$^{*}$ ($\tau{=}0.01$)~\cite{srinivasa2023cwcl} & 14.17 & 34.02 & 45.97 & 21.74 & 47.00 & 59.98 & 37.15 & 27.72 & 55.58 & 67.94 & 40.20 & 68.90 & 80.20 & 56.76 \\
                & CWCL$^{*}$ ($\tau{=}0.05$)~\cite{srinivasa2023cwcl} & 14.83 & 35.43 & 47.31 & 23.50 & 47.54 & 59.96 & 38.09 & 28.70 & 57.30 & 68.98 & 41.40 & 69.80 & 79.70 & 57.65 \\
                & MM-TS~\cite{sheludzko2026mm} & 21.20 & -- & -- & 28.40 & -- & -- & -- & 41.50 & -- & -- & \underline{54.30} & -- & -- & -- \\
                \rowcolor{oursbg}
                & TeMo (ours) & \textbf{23.28} & \textbf{47.48} & \textbf{59.62} & \textbf{30.36} & \textbf{56.82} & \textbf{67.92} & \textbf{47.58} & \textbf{44.16} & \textbf{73.26} & \textbf{81.84} & \textbf{56.00} & \textbf{82.70} & \textbf{90.80} & \textbf{71.46} \\
            \midrule
            \multirow{7}{*}{ViT-B/16}
                & InfoNCE & \underline{21.88} & \underline{45.60} & \underline{57.35} & \textbf{28.98} & \textbf{55.62} & \textbf{67.76} & \underline{46.20} & \underline{42.54} & \underline{69.00} & 77.74 & \underline{54.10} & \underline{81.70} & \underline{89.40} & \underline{69.08} \\
                & TS$^{*}$~\cite{kukleva2023temperature} & 21.45 & 44.31 & 55.61 & 27.32 & 53.14 & 65.88 & 44.62 & 41.02 & 68.86 & 78.68 & 49.90 & 80.90 & 88.60 & 67.99 \\
                & DySTreSS$^{*}$~\cite{manna2025dynamically} & 20.67 & 43.91 & 56.01 & 26.26 & 51.40 & 64.50 & 43.79 & 39.36 & 66.40 & 76.72 & 49.70 & 78.30 & 87.50 & 66.33 \\
                & DySTreSS Shifted$^{*}$~\cite{manna2025dynamically} & 21.33 & 44.80 & 56.82 & 25.40 & 52.20 & 64.66 & 44.20 & 40.74 & 68.64 & \underline{78.90} & 49.90 & 77.40 & 85.50 & 66.85 \\
                & CWCL$^{*}$ ($\tau{=}0.01$)~\cite{srinivasa2023cwcl} & 17.90 & 39.87 & 52.02 & 26.12 & 51.94 & 64.98 & 42.14 & 36.62 & 64.98 & 74.86 & 49.00 & 78.40 & 86.90 & 65.13 \\
                & CWCL$^{*}$ ($\tau{=}0.05$)~\cite{srinivasa2023cwcl} & 18.84 & 41.12 & 53.59 & 28.44 & 54.48 & 66.80 & 43.88 & 37.90 & 67.06 & 77.56 & 48.10 & 79.00 & 88.20 & 66.30 \\
                \rowcolor{oursbg}
                & TeMo (ours) & \textbf{22.87} & \textbf{46.66} & \textbf{58.88} & \underline{28.80} & \underline{55.48} & \underline{67.66} & \textbf{46.73} & \textbf{45.28} & \textbf{73.14} & \textbf{81.58} & \textbf{55.60} & \textbf{82.60} & \textbf{89.50} & \textbf{71.28} \\
            \bottomrule
        \end{tabular}
    }
   \caption{\textbf{Extended retrieval comparison} (extension of
    Table~\ref{tab:sota_retrieval_and_zs_classification}). Retrieval accuracy (\%)
    on MSCOCO and Flickr30k, reporting R@1/5/10 in both directions and R-Mean, the
    average over image-to-text and text-to-image recall. DySTreSS$^{*}$, DySTreSS
    Shifted$^{*}$, and CWCL$^{*}$ denote our multimodal
    adaptations/implementations of~\cite{manna2025dynamically}
    and~\cite{srinivasa2023cwcl}.}
    \label{tab:sota_retrieval_extended}
\end{table*}

\begin{table*}[h]
    \centering
    \small
    \setlength{\tabcolsep}{4pt}
    \renewcommand{\arraystretch}{1.1}
        \begin{tabular}{lll ccc ccc ccc}
            \toprule
            \multirow{2}{*}{\textbf{Dataset}} & \multirow{2}{*}{\textbf{Backbone}} & \multirow{2}{*}{\textbf{Method}}
                & \multicolumn{3}{c}{\textbf{CIFAR10}}
                & \multicolumn{3}{c}{\textbf{CIFAR100}}
                & \multicolumn{3}{c}{\textbf{IN-1k}} \\
            \cmidrule(lr){4-6}\cmidrule(lr){7-9}\cmidrule(lr){10-12}
            & & & Top-1 & Top-3 & Top-5 & Top-1 & Top-3 & Top-5 & Top-1 & Top-3 & Top-5 \\
            \midrule
            \multirow[t]{6}{*}{\textbf{CC3M}} & \multirow[t]{3}{*}{RN50}
                & InfoNCE & \underline{53.78} & 81.76 & 90.38 & 25.08 & 42.09 & 50.40 & 30.70 & 43.31 & 48.34 \\
                & & TS$^{*}$~\cite{kukleva2023temperature} & 51.61 & \underline{82.03} & \underline{92.08} & \underline{28.53} & \underline{45.21} & \underline{52.57} & \underline{31.86} & \underline{45.25} & \underline{50.38} \\
                \rowcolor{oursbg}
                & & TeMo (ours) & \textbf{64.04} & \textbf{89.85} & \textbf{95.99} & \textbf{37.52} & \textbf{57.48} & \textbf{65.97} & \textbf{37.47} & \textbf{50.98} & \textbf{55.93} \\
            \cmidrule(lr){2-12}
            & \multirow[t]{3}{*}{ViT-B/16}
                & InfoNCE & 70.55 & 90.21 & 95.50 & 44.53 & 61.41 & 68.13 & 28.00 & 40.23 & 45.15 \\
                & & TS$^{*}$~\cite{kukleva2023temperature} & \underline{72.02} & \underline{91.65} & \underline{96.03} & \underline{47.13} & \underline{65.45} & \underline{72.17} & \underline{29.01} & \underline{42.17} & \underline{47.20} \\
                \rowcolor{oursbg}
                & & TeMo (ours) & \textbf{81.54} & \textbf{94.47} & \textbf{98.08} & \textbf{53.62} & \textbf{70.39} & \textbf{76.30} & \textbf{34.79} & \textbf{48.32} & \textbf{53.53} \\
            \midrule
            \multirow[t]{4}{*}{\textbf{CC12M}} & \multirow[t]{4}{*}{ViT-B/16}
                & InfoNCE & 65.18 & 89.40 & 96.22 & 37.12 & 57.34 & 65.42 & 35.48 & 54.80 & 62.54 \\
                & & TS$^{*}$~\cite{kukleva2023temperature} & \underline{71.31} & \underline{91.28} & \underline{96.31} & 39.29 & 60.69 & 69.60 & 38.91 & 59.02 & \underline{67.27} \\
                & & SLIP~\cite{mu2021slip} & 68.50 & 89.27 & 94.59 & \underline{45.08} & \underline{66.38} & \textbf{74.57} & \underline{39.77} & \underline{63.05} & 67.08 \\
                \rowcolor{oursbg}
                & & TeMo (ours) & \textbf{78.63} & \textbf{92.42} & \textbf{96.42} & \textbf{45.24} & \textbf{66.49} & \underline{74.36} & \textbf{41.76} & \textbf{63.05} & \textbf{71.41} \\
            \bottomrule
        \end{tabular}
    \caption{\textbf{Extended zero-shot classification} (extension of
    Tables~\ref{tab:sota_retrieval_and_zs_classification}
    and~\ref{tab:cc12m_cc3m_zs}). Top-K accuracy (\%) on CIFAR10, CIFAR100, and
    ImageNet-1k for $K\in\{1,3,5\}$, with models pretrained on CC3M and CC12M. TS$^{*}$ is our
    multimodal adaptation of Temperature Schedules~\cite{kukleva2023temperature}.}
    \label{tab:sota_zshot_extended}
\end{table*}

\section{Ablations on Loss Components}
We provide the full component ablation on both evaluation tasks.
Table~\ref{tab:loss_components_extended} extends the retrieval ablation with
R@1, R@5, R@10, and R-Mean on MSCOCO and Flickr30k, and
Table~\ref{tab:loss_components_zsh_summary} reports the same configurations on
zero-shot Top-1 classification across the 17 CLIP Benchmark datasets. The
results confirm the trend from the main paper: adding each component improves
performance, and the full model, which combines the modulated multimodal and
unimodal losses with the progressive scheduler, performs best on the large
majority of metrics and datasets.

\begin{table*}[h]
    \centering
    \setlength{\tabcolsep}{4pt}
    \renewcommand{\arraystretch}{1.1}
    \resizebox{\textwidth}{!}{%
        \begin{tabular}{cccc ccc ccc c ccc ccc c}
            \toprule
            \multirow{2}{*}{\textbf{Base}} & \multirow{2}{*}{\textbf{Mod.}} & \multirow{2}{*}{\textbf{U.L.}} & \multirow{2}{*}{\textbf{Sch.}}
                & \multicolumn{3}{c}{\textbf{MSCOCO IR}} & \multicolumn{3}{c}{\textbf{MSCOCO TR}} & \multirow{2}{*}{\textbf{R-Mean}}
                & \multicolumn{3}{c}{\textbf{Flickr30k IR}} & \multicolumn{3}{c}{\textbf{Flickr30k TR}} & \multirow{2}{*}{\textbf{R-Mean}} \\
            \cmidrule(lr){5-7}\cmidrule(lr){8-10}\cmidrule(lr){12-14}\cmidrule(lr){15-17}
            & & & & R@1 & R@5 & R@10 & R@1 & R@5 & R@10 & & R@1 & R@5 & R@10 & R@1 & R@5 & R@10 & \\
            \midrule
            \checkmark & -- & -- & -- & 21.64 & 45.77 & 57.47 & 28.60 & 53.90 & 66.56 & 45.66 & 42.12 & 68.84 & 78.76 & 53.60 & 81.30 & 88.90 & 68.92 \\
            \midrule
            -- & \checkmark & -- & -- & 18.32 & 40.85 & 52.79 & 25.36 & 50.60 & 62.70 & 41.77 & 36.16 & 65.16 & 74.78 & 45.00 & 75.60 & 84.70 & 63.57 \\
            \checkmark & -- & \checkmark$^{\dagger}$ & -- & 20.65 & 43.81 & 55.85 & 25.44 & 51.86 & 64.02 & 43.61 & 39.16 & 67.46 & 78.06 & 50.10 & 78.00 & 86.30 & 66.51 \\
            \checkmark & \checkmark & -- & -- & 20.09 & 44.45 & 56.55 & 27.54 & 53.40 & 65.22 & 44.68 & 40.12 & 67.32 & 77.24 & 52.20 & 80.90 & 87.70 & 67.58 \\
            \checkmark & \checkmark & -- & \checkmark & \underline{22.44} & 46.06 & 58.07 & \underline{29.92} & \textbf{57.00} & \underline{67.98} & \underline{46.91} & \underline{43.24} & 70.56 & 79.40 & 54.50 & \underline{82.50} & \underline{89.80} & 70.00 \\
            \checkmark & \checkmark & \checkmark & -- & 21.90 & \underline{46.10} & \underline{58.25} & 29.38 & 56.54 & \textbf{68.62} & 46.80 & 42.84 & \underline{71.70} & \underline{80.94} & \underline{54.80} & 80.80 & 89.20 & \underline{70.05} \\
            \rowcolor{oursbg}
            \checkmark & \checkmark & \checkmark & \checkmark & \textbf{23.28} & \textbf{47.48} & \textbf{59.62} & \textbf{30.36} & \underline{56.82} & 67.92 & \textbf{47.58} & \textbf{44.16} & \textbf{73.26} & \textbf{81.84} & \textbf{56.00} & \textbf{82.70} & \textbf{90.80} & \textbf{71.46} \\
            \bottomrule
        \end{tabular}
    }
    \caption{\textbf{Extended loss-component ablation} (extension of
    Table~\ref{tab:loss_components_ablation}). Retrieval accuracy (\%) on MSCOCO
    and Flickr30k under different training configurations; our full model (last
    row) is highlighted, \textbf{best} per column in bold. Base: InfoNCE loss
    $\mathcal{L}_{\mathrm{MM}}$; Mod.: modulated multimodal loss
    $\mathcal{L}_{\mathrm{M\text{-}MM}}$; U.L.: additional unimodal losses;
    Sch.: progressive scheduler. $^{\dagger}$Unimodal objectives are applied
    without temperature modulation.}
    \label{tab:loss_components_extended}
\end{table*}

\begin{table*}[h]
    \centering
    \setlength{\tabcolsep}{3pt}
    \renewcommand{\arraystretch}{1.1}
    \resizebox{\textwidth}{!}{%
        \begin{tabular}{cccc *{17}{c} c}
            \toprule
            \multirow{2}{*}{\textbf{Base}} & \multirow{2}{*}{\textbf{Mod.}} & \multirow{2}{*}{\textbf{U.L.}} & \multirow{2}{*}{\textbf{Sch.}}
            & \multicolumn{17}{c}{\textbf{Zero-shot classification (Top-1 Acc.\ \%)}} & \multirow{2}{*}{\textbf{Avg}} \\
            \cmidrule(lr){5-21}
            & & & & C10 & C100 & C211 & DTD & ESAT & GTSRB & FGVCA & IN-S & IN-A & IN-R & IN-O & FLO & PETS & PCAM & MNIST & F101 & IN1k & \\
            \midrule
            \checkmark & -- & -- & -- & 53.78 & 25.08 & 1.28 & 24.20 & 17.34 & 8.11 & 1.14 & 19.19 & 7.12 & 39.01 & 37.20 & 14.34 & 21.32 & 49.84 & \textbf{13.30} & 20.29 & 30.70 & 22.54 \\
            \midrule
            -- & \checkmark & -- & -- & 50.85 & 29.08 & 1.08 & 19.41 & 16.04 & 7.73 & 0.87 & 19.18 & 8.16 & 38.78 & 36.95 & 12.31 & 16.89 & 50.69 & \underline{11.50} & 20.61 & 31.10 & 21.84 \\
            \checkmark & -- & \checkmark$^{\dagger}$ & -- & \underline{61.96} & 34.22 & \underline{1.58} & 23.35 & \textbf{29.73} & 6.48 & 1.05 & 21.11 & 7.89 & 39.68 & 41.70 & 11.46 & 17.71 & 57.24 & 6.03 & 17.74 & 36.74 & 24.45 \\
            \checkmark & \checkmark & -- & -- & 51.64 & 29.06 & 1.10 & 21.75 & 12.38 & 8.14 & \underline{1.29} & 19.90 & 7.69 & 39.18 & 39.25 & \underline{14.57} & 20.85 & 52.29 & 10.34 & 19.18 & 31.69 & 22.37 \\
            \checkmark & \checkmark & -- & \checkmark & 49.28 & 26.27 & 1.34 & 23.56 & 22.90 & 8.76 & 1.17 & 20.91 & 8.45 & 42.01 & 39.70 & \textbf{14.77} & \underline{21.83} & 54.94 & 10.01 & 20.67 & 32.94 & 23.50 \\
            \checkmark & \checkmark & \checkmark & -- & 60.92 & \underline{34.90} & 1.54 & \underline{26.22} & 16.26 & \textbf{9.60} & \textbf{1.71} & \underline{24.17} & \underline{9.19} & \underline{44.01} & \underline{45.40} & 12.93 & 18.42 & \textbf{61.81} & 8.61 & \underline{20.99} & \underline{37.04} & \underline{25.51} \\
            \rowcolor{oursbg}
            \checkmark & \checkmark & \checkmark & \checkmark & \textbf{64.04} & \textbf{37.52} & \textbf{1.68} & \textbf{27.29} & \underline{25.61} & \underline{9.14} & 1.08 & \textbf{25.17} & \textbf{9.81} & \textbf{44.42} & \textbf{46.55} & 13.04 & \textbf{21.91} & \underline{61.51} & 10.50 & \textbf{22.00} & \textbf{37.47} & \textbf{26.98} \\
            \bottomrule
        \end{tabular}
    }
    \caption{\textbf{Loss-component ablation on zero-shot classification.} Top-1
    accuracy (\%) across 17 datasets for the configurations of
    Table~\ref{tab:loss_components_extended}; \textbf{Avg} is the mean over the
    17 datasets. Base: InfoNCE $\mathcal{L}_{\mathrm{MM}}$; Mod.: modulated
    multimodal loss $\mathcal{L}_{\mathrm{M\text{-}MM}}$; U.L.: additional
    unimodal losses; Sch.: scheduler. $^{\dagger}$Unimodal objectives are
    applied without temperature modulation. Datasets: C10/C100 = CIFAR10/100,
    C211 = Country211, DTD = Describable Textures, ESAT = EuroSAT, GTSRB =
    Traffic Signs, FGVCA = FGVC-Aircraft, IN-S/A/R/O = ImageNet-Sketch/A/R/O,
    FLO = Flowers-102, PETS = Oxford-IIIT Pets, PCAM = PatchCamelyon, MNIST =
    MNIST, F101 = Food-101, IN1k = ImageNet-1k.}
    \label{tab:loss_components_zsh_summary}
\end{table*}

\section{$k$-NN Evaluation on Visual Representations}
Table~\ref{tab:knn_evaluation} reports $k$-NN accuracy on the visual
embeddings, the tabular counterpart of Figure~\ref{fig:modality_gap_and_knn}, extended with $k$-NN@10 alongside $k$-NN@1. Consistent with the main
paper, the unimodal losses are the component that drives this metric: every
configuration that includes them improves the visual neighborhood structure
markedly over those that do not, raising CIFAR10 $k$-NN@1 from around 73\% to
roughly 80\%. TeMo attains the best accuracy on three of the four metrics and
remains competitive on the fourth, confirming that its gains extend to the
geometry of the visual space and not only to cross-modal alignment.

\begin{table*}[h]
    \centering
    \setlength{\tabcolsep}{6pt}
    \renewcommand{\arraystretch}{1.1}
    \begin{tabular}{cccc cc cc}
        \toprule
        \multirow{2}{*}{\textbf{Base}} & \multirow{2}{*}{\textbf{Mod.}} & \multirow{2}{*}{\textbf{U.L.}} & \multirow{2}{*}{\textbf{Sch.}}
            & \multicolumn{2}{c}{\textbf{CIFAR10}}
            & \multicolumn{2}{c}{\textbf{CIFAR100}} \\
        \cmidrule(lr){5-6}\cmidrule(lr){7-8}
        & & & & $k$-NN@1 & $k$-NN@10 & $k$-NN@1 & $k$-NN@10 \\
        \midrule
        \checkmark & -- & -- & -- & 73.40 & 77.29 & 50.59 & 53.27 \\
        \midrule
        -- & \checkmark & -- & -- & 68.66 & 72.98 & 44.72 & 48.61 \\
        \checkmark & -- & \checkmark$^{\dagger}$ & -- & \underline{80.69} & \underline{83.88} & \underline{56.40} & \textbf{60.83} \\
        \checkmark & \checkmark & -- & -- & 73.43 & 77.43 & 50.36 & 53.87 \\
        \checkmark & \checkmark & -- & \checkmark & 72.14 & 76.13 & 49.44 & 52.55 \\
        \checkmark & \checkmark & \checkmark & -- & 80.27 & 83.69 & 56.18 & 59.68 \\
        \rowcolor{oursbg}
        \checkmark & \checkmark & \checkmark & \checkmark & \textbf{80.70} & \textbf{83.89} & \textbf{57.04} & \underline{60.65} \\
        \bottomrule
    \end{tabular}
    \caption{\textbf{$k$-NN evaluation on visual representations} (tabular
    version of Figure~\ref{fig:modality_gap_and_knn}). $k$-NN@1 and
    $k$-NN@10 accuracy (\%) on CIFAR10 and CIFAR100 under different training
    configurations. Base: InfoNCE loss $\mathcal{L}_{\mathrm{MM}}$; Mod.:
    modulated multimodal loss $\mathcal{L}_{\mathrm{M\text{-}MM}}$; U.L.:
    additional unimodal losses; Sch.: scheduler. $^{\dagger}$Unimodal
    objectives are applied without temperature modulation.}
    \label{tab:knn_evaluation}
\end{table*}

\section{Effect of the Similarity Source}
We ask whether the temperature modulation benefits from similarity matrices
produced by external expert encoders rather than by the model itself. For
images, we extract patch-level features with DINOv2-Small~\cite{oquab2023dinov2}
and average them into a global embedding per image, from which we compute the
image-to-image (I2I) similarity matrix. For text, we obtain sentence embeddings
from all-roberta-large-v1~\cite{reimers2019sentence} and use them for the
text-to-text (T2T) similarity matrix. We evaluate three variants: modulating
with the I2I expert only, the T2T expert only, and both together.

As shown in Tables~\ref{tab:retrieval_experts}
and~\ref{tab:zs-classification-experts}, expert similarities improve over the
InfoNCE baseline on most metrics, confirming that similarity-conditioned
modulation is beneficial regardless of its source. All expert variants are
nonetheless outperformed by TeMo, which derives its similarities directly from
the model's own embeddings. TeMo thus achieves stronger results without any
external encoder or added computation, underscoring its effectiveness and
efficiency as a self-contained approach.

\begin{table*}[ht]
    \centering
    \setlength{\tabcolsep}{4pt}
    \renewcommand{\arraystretch}{1.1}
    \resizebox{\textwidth}{!}{%
        \begin{tabular}{l ccc ccc c ccc ccc c}
            \toprule
            \multirow{2}{*}{\textbf{Method}}
                & \multicolumn{3}{c}{\textbf{MSCOCO IR}} & \multicolumn{3}{c}{\textbf{MSCOCO TR}} & \multirow{2}{*}{\textbf{R-Mean}}
                & \multicolumn{3}{c}{\textbf{Flickr30k IR}} & \multicolumn{3}{c}{\textbf{Flickr30k TR}} & \multirow{2}{*}{\textbf{R-Mean}} \\
            \cmidrule(lr){2-4}\cmidrule(lr){5-7}\cmidrule(lr){9-11}\cmidrule(lr){12-14}
            & R@1 & R@5 & R@10 & R@1 & R@5 & R@10 & & R@1 & R@5 & R@10 & R@1 & R@5 & R@10 & \\
            \midrule
            InfoNCE & 21.64 & 45.77 & 57.47 & 28.60 & 53.90 & 66.56 & 45.66 & 42.12 & 68.84 & 78.76 & 53.60 & 81.30 & 88.90 & 68.92 \\
            \midrule
            Text Expert & \underline{22.01} & \underline{45.84} & \underline{57.63} & 28.76 & 55.48 & 67.22 & 46.16 & \underline{42.46} & \underline{69.64} & \underline{79.12} & 53.10 & \underline{81.50} & 88.90 & \underline{69.12} \\
            Vision Expert & 21.87 & 45.47 & 57.30 & 28.60 & \underline{56.00} & 67.38 & 46.10 & 41.58 & 68.88 & 78.24 & 53.20 & 80.30 & 87.40 & 68.71 \\
            Text \& Vision Expert & 21.93 & 45.55 & 57.52 & \underline{29.56} & 55.96 & \underline{67.46} & \underline{46.33} & 41.00 & 69.12 & 78.82 & \underline{54.10} & 80.50 & \underline{89.00} & 68.75 \\
            \rowcolor{oursbg}
            TeMo (ours) & \textbf{23.28} & \textbf{47.48} & \textbf{59.62} & \textbf{30.36} & \textbf{56.82} & \textbf{67.92} & \textbf{47.58} & \textbf{44.16} & \textbf{73.26} & \textbf{81.84} & \textbf{56.00} & \textbf{82.70} & \textbf{90.80} & \textbf{71.46} \\
            \bottomrule
        \end{tabular}
    }
    \caption{\textbf{Retrieval results with expert-informed similarities.}
    Retrieval performance (\%) on MSCOCO and Flickr30k using expert-informed
    similarity matrices for temperature modulation. Our method is highlighted;
    \textbf{best} per column in bold, \underline{second best} underlined.}
    \label{tab:retrieval_experts}
\end{table*}

\begin{table}[b]
    \centering
    \setlength{\tabcolsep}{3pt}
    \renewcommand{\arraystretch}{1.1}
    \resizebox{\columnwidth}{!}{%
        \begin{tabular}{l ccc ccc}
            \toprule
            \multirow{2}{*}{\textbf{Method}}
                & \multicolumn{3}{c}{\textbf{CIFAR10}} & \multicolumn{3}{c}{\textbf{CIFAR100}} \\
            \cmidrule(lr){2-4}\cmidrule(lr){5-7}
            & Top-1 & Top-3 & Top-5 & Top-1 & Top-3 & Top-5 \\
            \midrule
            InfoNCE & 53.78 & 81.76 & 90.38 & 25.08 & 42.09 & 50.40 \\
            \midrule
            Text Expert & 51.09 & 79.45 & 88.51 & 24.56 & 41.00 & 49.67 \\
            Vision Expert & 54.12 & \underline{83.90} & \underline{93.67} & \underline{30.09} & \underline{47.16} & \underline{56.01} \\
            Text \& Vision Expert & \underline{54.95} & 83.00 & 93.61 & 26.77 & 43.45 & 52.00 \\
            \rowcolor{oursbg}
            TeMo (ours) & \textbf{64.04} & \textbf{89.85} & \textbf{95.99} & \textbf{37.52} & \textbf{57.48} & \textbf{65.97} \\
            \bottomrule
        \end{tabular}
    }
    \caption{\textbf{Zero-shot classification with expert-derived similarities.}
    Top-K accuracy (\%) on CIFAR10 and CIFAR100 when the temperature modulation
    uses expert-derived similarity matrices. Our method is highlighted;
    \textbf{best} per column in bold, \underline{second best} underlined.}
    \label{tab:zs-classification-experts}
\end{table}

\section{Robustness to False Positives}
\label{sec:robustness_agains_fn_sec}
\begin{figure*}[h]
    \centering
    \includegraphics[width=0.7\linewidth]{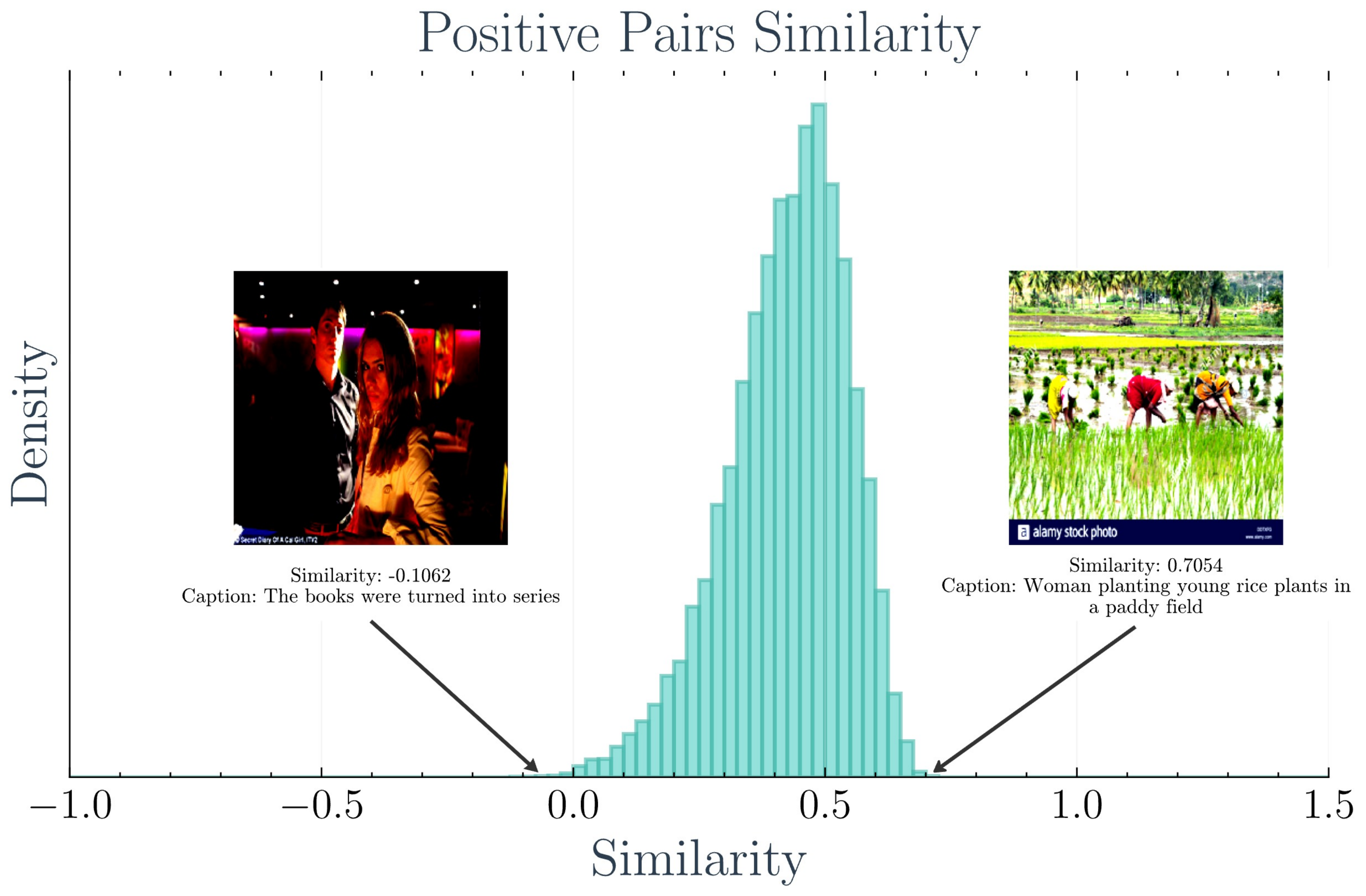}
    \caption{\textbf{Similarity distribution of positive pairs.} TeMo separates
    semantically aligned pairs from mismatched ones. Additional examples are
    provided in Figure~\ref{fig:false_positive_more_examples}.}
    \label{fig:false_positives}
\end{figure*}

\begin{figure*}[h]
    \centering
    \includegraphics[width=0.8\linewidth]{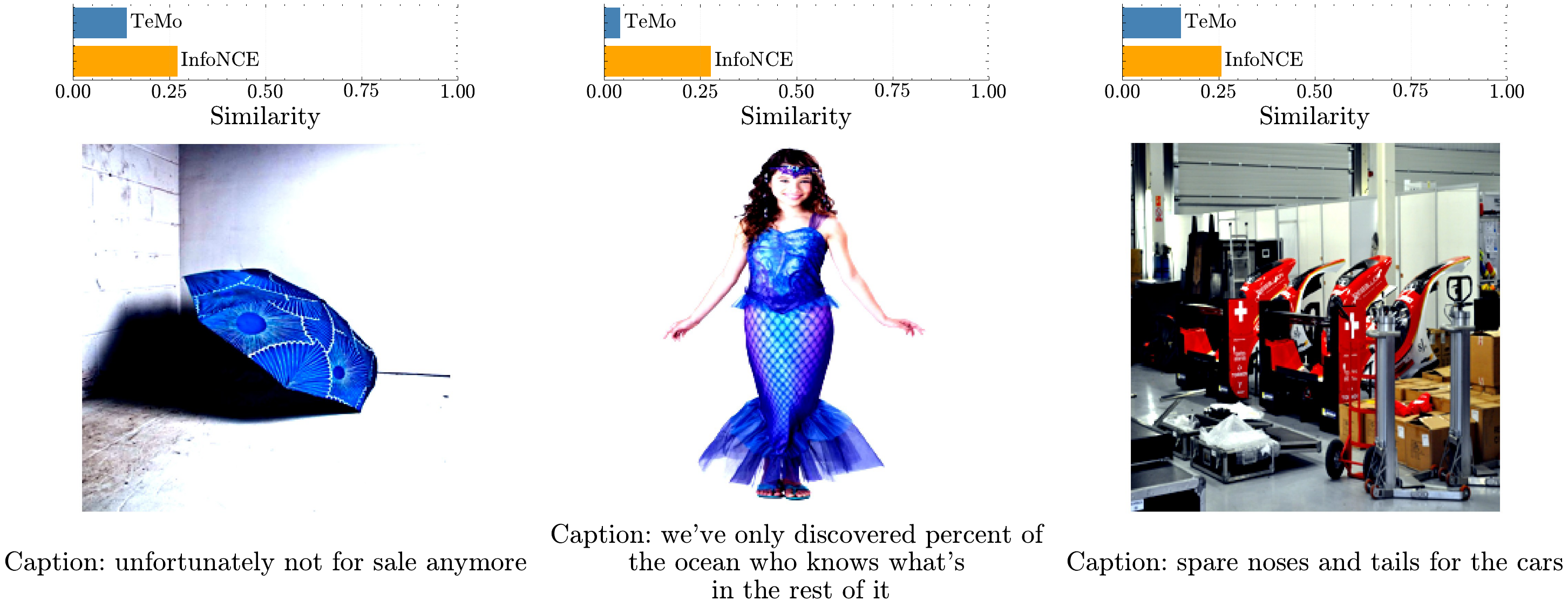}
    \caption{\textbf{Examples of false-positive image--caption pairs.} InfoNCE
    assigns high similarity to these mismatched examples, whereas
    TeMo assigns consistently lower scores, indicating improved discrimination
    and robustness to label noise.}
    \label{fig:false_positive_more_examples}
\end{figure*}

TeMo learns to down-weight false positives, i.e., image--caption pairs that are
labeled as matching in the dataset but are in fact mismatched.
Figures~\ref{fig:false_positives} and~\ref{fig:false_positive_more_examples}
illustrate this on the distribution of positive-pair similarities. At the
low-similarity end of Figure~\ref{fig:false_positives}, the captions do not
describe their images; these are genuine false positives, and TeMo assigns them
low scores. At the high-similarity end, the pairs are strongly aligned and are
scored accordingly. The standard deviation of the positive-pair scores is larger
for TeMo than for InfoNCE, indicating that TeMo reinforces truly meaningful
associations while suppressing the influence of noisy labels.

\section{Connection to Alignment and Uniformity}
TeMo can be understood as a dynamic mechanism that balances the alignment
and uniformity of the representation by adaptively modulating the temperature
$\tau$. Following~\cite{wang2021understanding}, a low temperature sharpens the
gradients that repel dissimilar pairs and thus promotes global uniformity,
whereas a high temperature relaxes the penalty on highly similar pairs and thus
favors alignment. TeMo applies this trade-off at the level of individual pairs:
dissimilar pairs receive low temperatures to spread the space, and similar
pairs receive high temperatures to preserve their semantic proximity. The
progressive scheduler turns this into a coarse-to-fine process. Early in
training, the temperature is fixed and low, expanding the embedding space and
preventing representation collapse; later, per-pair modulation is activated to
refine local neighborhoods and consolidate the semantic clusters formed
earlier, without over-separating related concepts. Our ablations support this
ordering: enforcing modulation before a uniform structure is established
degrades performance (Table~\ref{tab:loss_components_ablation}, row~b),
confirming the need to progress from uniformity to alignment.

\section{Effect on the Smoothness of the Objective}
Temperature modulation also acts as a data-dependent regularizer that
smooths the contrastive objective. With a fixed low temperature, high-similarity
negative pairs produce very sharp gradients that form steep regions in the loss
landscape and can destabilize training. Because TeMo raises the temperature as
similarity grows, it locally softens the softmax over these high-similarity
pairs, dampening their gradient magnitude and preventing the explosive updates
that hard negatives would otherwise cause.

\section{Limitations}
Our experiments are constrained by both batch size and pretraining dataset scale. Due to limited resources, we train with a maximum batch size of 4096 and only pretrain on CC3M and CC12M, with the latter being the largest dataset we use. In contrast, many recent works rely on significantly larger batch sizes (e.g., LaCLIP and LaSLIP~\cite{fan2023improving} at 8192, and large-scale models such as EvaCLIP~\cite{sun2023eva}, SigLIP~\cite{zhai2023sigmoid}, SigLIP2~\cite{tschannen2025siglip}, and TULIP~\cite{Tang_2025_ICCV} with 30k--170k) as well as much larger and more diverse pretraining datasets (e.g., YFCC15M~\cite{gu2024rwkv}, LAION~\cite{schuhmann2022laion}, or merged multi-billion scale datasets). Since performance is highly sensitive to both batch size and data scale, our results with TeMo should be viewed as competitive under constrained settings, rather than directly comparable to large-batch, large-data methods.

\section{LLM Usage}
A large language model was used to improve the clarity and readability of the
paper. All edits were reviewed and approved by the authors.

\end{document}